\documentclass[lettersize,journal]{IEEEtran}
\usepackage{amsmath,amsfonts}
\usepackage{algorithmic}
\usepackage{algorithm}
\usepackage{array}
\usepackage[caption=false,font=normalsize,labelfont=sf,textfont=sf]{subfig}
\usepackage{textcomp}
\usepackage{stfloats}
\usepackage{url}
\usepackage{verbatim}
\usepackage{graphicx}
\usepackage{cite}
\newcommand{\eg}{e.g.}
\newcommand{\ie}{i.e.}
\newcommand{\etal}{et al.}
\newcommand{\etc}{etc}
\usepackage{xcolor}
\usepackage{multirow}
\usepackage{booktabs}
\usepackage{makecell}

\usepackage[backref=true, colorlinks=true, linkcolor=black, citecolor=black]{hyperref}
\usepackage{cleveref}
\usepackage{diagbox}

\begin{document}

\title{UniAV: Unified Audio-Visual Perception for Multi-Task  Video Event Localization}

\author{Tiantian Geng,~\IEEEmembership{} 
Teng Wang,~\IEEEmembership{}
Jinming Duan,~\IEEEmembership{Member, IEEE},
Yanfu Zhang,~\IEEEmembership{}
Weili Guan,~\IEEEmembership{Member, IEEE},\\
Feng Zheng,~\IEEEmembership{Member, IEEE},
and Ling Shao,~\IEEEmembership{Fellow, IEEE}

\thanks{This work was supported in part by the National Key Research and Development Program of China under Grant 2024YFE0203100, in part by the National Outstanding Youth Science Fund Project of National Natural Science Foundation of China under Grant 62122035, and in part by the British Heart Foundation Manchester Research Excellence Award under Grant RE/24/130017.
\textit{(Corresponding authors: Feng Zheng and Jinming Duan.)}}
\thanks{Tiantian Geng is with the Department of Computer Science and Engineering, Southern University of Science and Technology, Shenzhen 518055, China, and also with the School of Computer Science, University of Birmingham, B15 2TT Birmingham, U.K. (e-mail: gengtiantian97@gmail.com).}
\thanks{Teng Wang is with the Department of Computer Science and Engineering, Southern University of Science and Technology, Shenzhen 518055, China, and also with the Department of Computer Science, University of Hong Kong, Hong Kong (e-mail: tengwang@connect.hku.hk).}
\thanks{Jinming Duan is with the School of Computer Science, University of Birmingham, B15 2TT Birmingham, U.K., and also with the Division of Informatics, Imaging and Data Sciences, University of Manchester, M13 9PL Manchester, U.K. (e-mail: j.duan@bham.ac.uk, jinming.duan@manchester.ac.uk).}
\thanks{Yanfu Zhang is with William and Mary, Williamsburg, VA 23185 USA (e-mail: yaz91@pitt.edu).}
\thanks{Weili Guan is with the Harbin Institute of Technology, Shenzhen 518057, China (e-mail: honeyguan@gmail.com).}
\thanks{Feng Zheng is with the Department of Computer Science and Engineering, Southern University of Science and Technology, Shenzhen 518055, China (email: f.zheng@ieee.org).}
\thanks{Ling Shao is with the UCAS-Terminus AI Lab, University of Chinese Academy of Sciences, Beijing 101408, China (e-mail: ling.shao@ieee.org).}
}

\maketitle

\begin{abstract}
Video event localization tasks include temporal action localization (TAL), sound event detection (SED) and audio-visual event localization (AVEL).
Existing methods tend to over-specialize on individual tasks, neglecting the equal importance of these different events for a complete understanding of video content.
In this work, we aim to develop a unified framework to solve TAL, SED and AVEL tasks together to facilitate holistic video understanding.
However, it is challenging since different tasks emphasize distinct event characteristics and there are substantial disparities in existing task-specific datasets (size/domain/duration). It leads to unsatisfactory results when applying a naive multi-task strategy.
To tackle the problem, we introduce UniAV, a Unified Audio-Visual perception network to effectively
learn and share mutually beneficial knowledge across tasks and modalities.
Concretely, we propose a unified audio-visual encoder to derive generic representations from multiple temporal scales for videos from all tasks. 
Meanwhile, task-specific experts are designed to capture the unique knowledge specific to each task. 
Besides, instead of using separate prediction heads, we develop a novel unified language-aware classifier by utilizing semantic-aligned task prompts, enabling our model to flexibly localize various instances across tasks with an impressive open-set ability to localize novel categories. 
Extensive experiments demonstrate that UniAV, with its unified architecture, significantly outperforms both single-task models and the naive multi-task baseline across all three tasks. 
It achieves superior or on-par performances compared to the state-of-the-art task-specific methods on ActivityNet 1.3, DESED and UnAV-100 benchmarks.

\end{abstract}


\maketitle
\section{Introduction}
\label{sec:intro}
\begin{figure}[!t]
  \centering
  \setlength{\abovecaptionskip}{0.5mm}
   \includegraphics[width=1.0\linewidth]{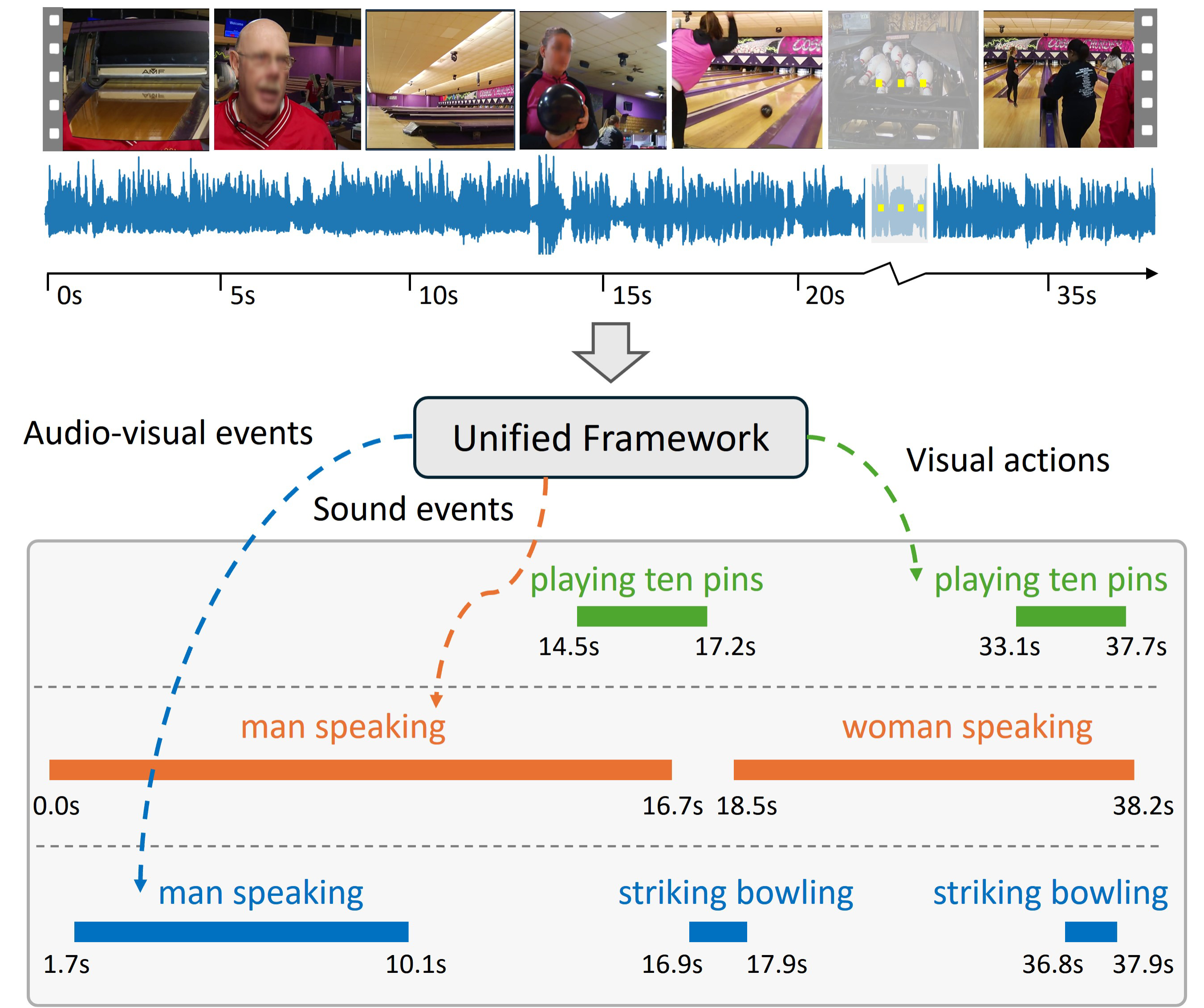}
   \caption{We propose a unified audio-visual perception framework to localize all three kinds of instances in untrimmed videos, including visual actions, sound events and audio-visual events. All these instances equally contribute to the comprehensive understanding of video content. }
   \label{fig:fig1}
\end{figure}

\IEEEPARstart{W}{ith} the explosion of video content due to social networks and digital cameras, video understanding~\cite{wang2018temporal,lin2018bsn, geng2022spatial,wang2022semantic,zeng2021graph} continues to be one of the essential research domains in computer vision.
Videos recorded in natural scenes are always untrimmed and comprise both visual and audio modalities. 
They usually cover multiple instances of interest, including visible-only actions, audible-only sound events as well as audio-visual events~\cite{tian2018audio} that are both audible and visible at the same time. 
Localizing these various instances automatically using audio and visual signals in a video has wide practical applications including intelligent surveillance, human behavior analysis and human-robot interaction.
As illustrated in Fig.~\ref{fig:fig1}, we can discern the visual action of ``playing ten pins'', the audio-visual events of ``striking bowling'', ``man speaking'', and also the out-of-screen narration of ``man/woman speaking''.
All these events are equally crucial, jointly contributing to the overall understanding of video content.

However, existing video localization approaches only concentrate on recognizing and detecting one type of these instances, involving the tasks of temporal action localization (TAL)~\cite{lin2019bmn,zeng2021graph,zhang2022actionformer,liu2022end,zhu2023contextloc++}, sound event detection (SED)~\cite{kim2022sound,xiao2023fmsg} and audio-visual event localization (AVEL)~\cite{tian2018audio,wu2019dual,wang2022semantic,zhou2022contrastive,geng2023dense}.
Despite convenience for some specific applications, such separate definitions bring the following drawbacks: 
1) Independent designs cause redundant parameters since recent localization models usually adopt similar architectures, \eg, transformer backbones in~\cite{zhang2022actionformer,liu2022end,geng2023dense}.
2) It hinders models from learning and sharing mutually beneficial knowledge between different tasks and modalities. 
For example, rich TAL data enables models to identify common instances, which can naturally assist AVEL and SED tasks.
Additionally, AVEL data allows models to learn corresponding audio representations for many visual actions in TAL and learn visual cues for sound events in SED, thereby facilitating improvements in both TAL and SED tasks.
Besides, some recent task-specific methods~\cite{bagchi2021hear,lee2020cross,hebbar2023dataset} have verified that integrating visual and audio modalities is beneficial for both TAL and SED tasks.

In this work, we aim to develop a unified framework to localize all three types of instances in untrimmed videos, solving three temporal video localization tasks (TAL, AVEL and SED) by a single unified model.
However, the main obstacles hindering this attempt lie in two aspects.
Firstly, different tasks emphasize different video characteristics and modalities.
TAL pays more attention to capturing temporal relationships of actions on the visual track, \eg, the visual movement of ``playing ten pins'' in Fig.~\ref{fig:fig1}.
SED is dedicated to the fine-grained understanding of audio signals where the events might be invisible, but still contain critical information, such as a narrating person and out-of-screen running cars.
In contrast, AVEL assigns equal importance to both auditory and visual cues to locate modality semantic-aligned snippets. For example, in Fig.~\ref{fig:fig1}, we can simultaneously see the image and hear the sound of a man speaking from 1.7s to 10.1s.
Secondly, the existing datasets for these tasks exhibit distinct properties with significant domain and duration gaps.
For example, ActivityNet~\cite{caba2015activitynet} for TAL task focuses on human daily activities, while UnAV-100~\cite{geng2023dense} for AVEL task and DESED~\cite{turpault:hal-02160855} for SED task contain events from other domains such as animals, nature and tools. 
Besides, the instance duration of different datasets varies greatly. For instance, over 50\% of sound events in DESED span less than one second, while the instances in ActivityNet and UnAV-100 are much longer, with the longest ones lasting over 7 minutes.
Therefore, unifying these three distinct tasks is challenging, and just naive multi-task training may lead to a significant decrease in performance.

To tackle the challenges, we introduce UniAV, a Unified Audio-Visual perception framework that enables the model to not only learn and share mutually beneficial knowledge across tasks and modalities but also capture the unique insights inherent to each task.
Specifically, we integrate TAL, SED and AVEL tasks into a cohesive framework from the following three aspects.
1) \textbf{Unified audio-visual encoding.} 
In order to unify diversity between the data from different tasks and obtain generic audio-visual representations, we first employ the audio and visual encoders from the large pre-trained model~\cite{wang2023one} to uniformly tokenize input videos from all tasks.
Then, the obtained embeddings are fed into an audio-visual pyramid transformer network, enabling the model to capture both very short as well as long instances by the thorough cross-modal fusion at multiple temporal scales.
2) \textbf{Task-specific experts.} Due to the substantial divergence of different tasks, learning distinct knowledge for each task is also critical.
Thus, we design task-specific expert layers in the transformer blocks to learn task-specific features for each task by adaptively switching to corresponding experts according to the input data.
3) \textbf{Unified language-aware classifier.} Datasets for different tasks pose their own category sets. Instead of using separate task-specific classification heads, we propose a unified language-aware classifier by tokenizing the class vocabularies with task-specific prompts using the pre-trained semantic-aligned text encoder~\cite{wang2023one}.
Benefiting from this new formulation, our model gains the remarkable flexibility to localize rich instance categories across tasks by simply changing task prompts, greatly alleviating the data scarcity of each task, especially for SED task, \ie, only 10 classes in DESED dataset. Furthermore, it enables our model to have an impressive open-vocabulary capability to localize novel instance categories, even across tasks during inference.

With the unified framework, UniAV can learn from diverse task-specific data and handle three video event localization tasks using a single model parameters.
Extensive experiments demonstrate that 
UniAV outperforms its single-task counterparts and the naive multi-task baseline by a large margin, and achieves superior or on-par performances compared to the latest task-specific models.
Besides, we find that multi-task joint training can be an effective pre-training step for the single-task models, leading to further performance gains and setting new state-of-the-art results across all three tasks,
\ie, ActivityNet 1.3~\cite{caba2015activitynet} ($36.2\%$ average mAP) for TAL, DESED~\cite{kim2022sound} ($61.1\%$ average mAP) for SED, and UnAV-100~\cite{geng2023dense} ($51.7\%$ average mAP) for AVEL. 

\noindent Our contributions can be summarized as follows:
\begin{itemize}
    \item 
    To the best of our knowledge, our UniAV is the first unified framework that solves temporal action localization, sound event detection and audio-visual event localization tasks within a single model, leading to a holistic understanding of video content in real-world scenarios. 
    \item  We propose a unified audio-visual encoding pipeline to address substantial data discrepancies across tasks, and meanwhile design task-specific experts to effectively capture distinct knowledge for each task.
    \item  We introduce a unified language-aware classifier with semantic-aligned task prompts, allowing the model to flexibly detect various instances across tasks and have an open-set localization capability during inference.
    \item Experimental results show that UniAV, with its unified architecture, significantly outperforms both single-task models and the naive multi-task baseline, achieving state-of-the-art performances across all three task benchmarks.
\end{itemize}
\section{Related Work}
\label{sec:related_work}

\subsection{Video Event Localization Tasks}
{\bf Temporal action localization (TAL)} aims to localize and classify action instances occurring in an untrimmed video. Supervised learning-based TAL can be categorized into two-stage and single-stage methods. 
Two-stage methods~\cite{lin2019bmn,lin2018bsn,xu2019two,bai2020boundary,xu2020g,zhu2023contextloc++} first generate action proposals with confidence scores, and then classify their corresponding action segments and refine the temporal boundaries.
By contrast, single-stage pipelines~\cite{lin2021learning,liu2022end,tan2021relaxed,zhang2022actionformer,shi2023tridet,kim2024te} simultaneously predict temporal boundaries and action categories for each instance.
Besides, some works explore new mechanisms to solve TAL. Yang \etal~\cite{yang2024adapting} propose adapters into short-term Vision transformer (ViT)~\cite{dosovitskiy2020image} for end-to-end TAL without feature extraction. Foo \etal~\cite{foo2024action} propose an image-diffusion framework to re-cast TAL as an image generation process.
These previous works only focus on temporal modeling within the visual modality (\eg, RGB frames and optical flow), ignoring the information in its corresponding audio track. 
In recent years, some works~\cite{bagchi2021hear,lee2020cross,lee2022leaky} attempted to utilize the audio modality in videos for TAL, and found it very helpful to detect the actions that have strong audio cues, thus boosting the model performance. 

{\bf Sound event detection (SED)} is a popular task in the audio signal processing community, which involves temporally detecting sound events in purely acoustic scenes. The DCASE Challenge~\cite{mesaros2017dcase} examines sound event detection in domestic environments as one of the challenge tasks~\cite{khandelwal2022fmsg,kim2022sound,xiao2023fmsg}.
However, sound events typically come with their corresponding visual information, \eg, the sound events of man speaking, playing guitar, and dog barking.
It has been verified that incorporating visual modality is beneficial for SED tasks in some recent works~\cite{hebbar2023dataset,boes2021audiovisual}.

{\bf Audio-visual event localization (AVEL)} aims to detect events that are simultaneously audible and visible in video content.
Tian \etal~\cite{tian2018audio} introduced the first AVEL dataset and proposed an audio-guided visual attention model for the task. {\color{black}Subsequent studies~\cite{wu2019dual,xu2020cross,xuan2020cross,duan2021audio,wang2022semantic,zhou2022contrastive} primarily concentrated on event information modeling and cross-modal fusion strategies, and  Zhou \etal~\cite{zhou2024towards} extended AVEL to a new open-vocabulary setting.}
However, these works formulate AVEL as a segment-level classification problem based on trimmed, short video clips. Each video clip only contains a single audio-visual event, which deviates from real-life scenarios involving diverse untrimmed videos.
To solve the problem, Geng \etal~\cite{geng2023dense} built the UnAV-100 dataset for localizing audio-visual events in untrimmed videos and proposed a model to recognize multiple events and regress their temporal boundaries in a single pass. 
Furthermore, few approaches for audio-visual video parsing~\cite{tian2020unified,lin2021exploring,mo2022multi,rachavarapu2023boosting} and multisensory temporal event localization~\cite{hou2024toward} strive to identify audio-only, visual-only and audio-visual events in videos. However, they are all segment-level classification methods with fragmented task definitions, and confined to weakly-supervised settings due to the lack of temporal annotations in videos during training.
Besides, the very limited instance categories, \ie, only 25 classes in LLP dataset~\cite{tian2020unified}, make the trained model incapable of detecting diverse instances in complex real-life scenarios.
In comparison, we develop a multi-task supervised framework that learns multi-modal event localization from large-scale individually collected datasets for TAL, SED and AVEL tasks with rich label vocabularies. Moreover, our model is proposal-based, capable of flexibly regressing and recognizing all temporal instances in untrimmed videos with various lengths.

\subsection{Multi-Task Learning}
Instead of separate training for each individual task, multi-task learning involves tackling multiple related tasks simultaneously, aiming to share and leverage knowledge across them.
For example, recent works~\cite{luo2020multi,li2021referring,su2023language}
proposed to jointly learn visual grounding tasks in a collaborative model.
Yan \etal~\cite{yan2022towards} presented Unicorn to solve four object tracking problems. 
Artacho \etal~\cite{artacho2021unipose+} introduced UniPose+ to unify 2D and 3D human pose estimation tasks by a single-pass multi-scale architecture.
Lu \etal~\cite{lu202012} undertook training across 12 vision-language tasks, with each task having its own task-specific prediction head. 
For video understanding tasks, UniVTG~\cite{lin2023univtg} stands out for its emphasis on unifying three temporal grounding tasks: moment retrieval, highlight detection, and video summarization, employing specific query types. 
Addressing episodic memory tasks, MINOTAUR~\cite{goyal2023minotaur} exhibits proficiency in handling three egocentric vision tasks~\cite{grauman2022ego4d} through a singular model.
{\color{black}Furthermore, recent multimodal large language models (MLLMs)~\cite{shu2023audio,cheng2024videollama,tang2024avicuna,guo2024trace,wu2024number,chowdhury2024meerkat} show promise in unified multi-task learning. For instance, VideoLLaMA~\cite{cheng2024videollama} handles video question answering and captioning tasks simultaneously, and Trace~\cite{guo2024trace} uniformly formats time-aware video tasks into text generation.
However, LLM-based methods consume much more computational resources and often underperform compared to non-LLM methods.}
In this work, we unify temporal action localization, sound event detection, and audio-visual event localization tasks for the first time to achieve diverse modality-aware instance localization in untrimmed videos for holistic video understanding. 
This is a distinct and pressing problem that was previously overlooked but holds important practical implications in real-world audio-visual scenarios.
\section{The UniAV Framework}
\begin{figure*}[!t]
  \centering
  \setlength{\abovecaptionskip}{0.5mm}
   \includegraphics[width=1.0\linewidth]{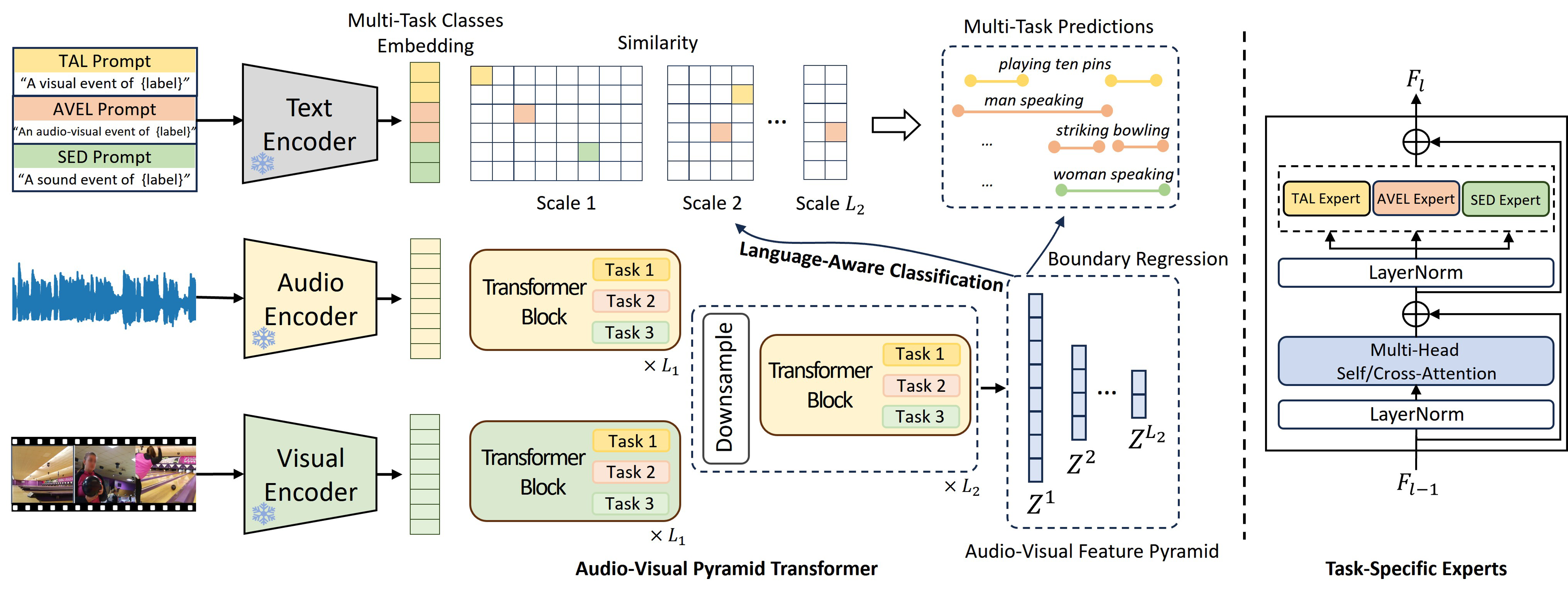}
   \caption{
   The overview of our unified framework. Given a visual and audio pair from an untrimmed video, we first tokenize them by a pair of pre-trained visual and audio encoders to obtain generic audio and visual representations.
   Then, the encoded features are fed into an audio-visual pyramid transformer, which consists of $L_{1}$ uni-modal and $L_{2}$ cross-modal transformer blocks for temporal relation modeling and audio-visual fusion at multiple temporal scales.    
   The task-specific experts inserted in transformer blocks learn distinct knowledge for each task, which is illustrated on the right side. 
   Besides, the categories of each task are encoded with task prompts to compute similarities with the obtained audio-visual feature pyramid, which is used to perform language-aware classification.  
   Finally, the model recognizes classes and regresses temporal boundaries for all types of instances occurring in the video. }
   \label{fig:framwork}
\end{figure*}

Our goal is to develop a unified framework to localize visual actions, sound events and audio-visual events in an untrimmed video.
To achieve this, we propose to unite three video localization tasks: TAL, SED and AVEL, leveraging inherent similarities among them.
An overview of the proposed framework is illustrated in Fig.~\ref{fig:framwork}.

{\bf Problem statement.}
Given an untrimmed video containing visual and audio tracks, our model aims to recognize the instance categories and locate the start/end timestamps of all modality-aware instances occurring in the video.
Different from the previous fragmented task definitions~\cite{zhang2022actionformer,tian2018audio,khandelwal2022fmsg}, we uniformly formulate all three tasks (TAL, SED and AVEL) as a joint classification and regression problem.

Formally, we first divide the input video into $T$ visual snippets $\{V_{t}\}_{t=1}^{T}$ and audio snippets $\{A_{t}\}_{t=1}^{T}$, where $T$ varies across videos. 
For the given task, the groundtruth of an input video is denoted as $Y=\{y_{n}=(t_{s,n}, t_{e,n}, c_{n})\}^{N}_{n=1}$, where $t_{s,n}$, $t_{e,n}$ are the start and end timestamps of $n$-th instance, $c_{n}\in\{1, \cdots, C\}$ is the instance category, and $N$ is the number of instances occurring in the video.
Following the setting~\cite{zhang2022actionformer, geng2023dense}, the model is expected to predict 
$\hat{Y} = \{\hat{y}_{t}=(d_{s,t}, d_{e,t}, p(c_{t}))\}_{t=1}^{T}$ for each moment during inference, where $p(c_{t})\in \mathbb{R}^{1\times C}$ is the probabilities of $C$ categories at moment $t$, $d_{s,t}$ and $d_{e, t}$ are the distances between the moment $t$ to the instance's start and end timestamp, respectively. $d_{s,t}$ and $d_{e,t}$ are only defined when an instance presents at moment $t$.
Therefore, the prediction results for the given task can be decoded from $\hat{y}_{t}=(d_{s,t}, d_{e,t}, p(c_{t}))$ by:
\begin{equation}
    c_{t} = {\arg\max}\, p(c_{t}),\quad t_{s,t}=t-d_{s,t},\quad t_{e,t}=t+d_{e,t}.
\label{equ1} 
\end{equation}

\subsection{Unified Audio-Visual Encoding}
\label{sec: encoding}
Audio and visual signals are equally crucial for AVEL~\cite{tian2018audio,geng2023dense}, and both serve as important complementary information for TAL and SED to boost the localization performances~\cite{bagchi2021hear,boes2021audiovisual,hebbar2023dataset}. Therefore, to solve these three tasks together using a single unified framework, we propose a unified audio-visual encoding pipeline to utilize both audio and visual modalities of all input videos for all tasks. The proposed pipeline consists of two encoding processes as follows:

{\bf Audio and visual generic representations.}
Individually collected datasets for different tasks exhibit significant variations in domains and scales. 
In order to unify representations of diverse data to minimize data discrepancies, we use a general model~\cite{wang2023one} pre-trained by aligning vision, audio and language modalities to extract video representations.
Specifically, the visual and audio encoders of the model~\cite{wang2023one} are utilized to tokenize visual and audio snippets of a given input video, respectively. Note that the parameters of the encoders are frozen during this process. 
Since the visual encoder just learned from image data while temporal information in videos is essential for TAL and AVEL tasks, the visual encoder fine-tuned on Kinetics-400~\cite{carreira2017quo} is also considered. 
After tokenization, we apply a linear projection layer to project the audio and visual representations into a shared embedding space. Then, the visual $E_{V} = \{e_{t}^{v}\}_{t=1}^{T} \in \mathbb{R}^{T\times D}$ and audio $E_{A} = \{e_{t}^{a}\}_{t=1}^{T} \in\mathbb{R}^{T\times D}$ embedding sequences can be obtained, where $D$ is the projected feature dimension.

{\bf Audio-visual pyramid transformer.}
We consider that there is often irrelevant information and noise in audio and video tracks that interfere with accurate localization for specific types of instances, such as off-screen voice/music for AVEL,  a visible dog without barking for SED. 
Besides, diverse modality-aware instances occurring in untrimmed videos might span various duration. For instance, the event duration ranges from 0.2 to 60 seconds in UnAV-100~\cite{geng2023dense} for AVEL.
Therefore, an effective cross-modal fusion strategy is essential for solving all the three tasks. Here, we apply an audio-visual pyramid transformer inspired by~\cite{zhang2022actionformer,geng2023dense}.

Specifically, the tokenized visual $E_{V}$ and audio $E_{A}$ embedding sequences added with its position embeddings $E_{pos}\in \mathbb{R}^{T \times D}$ first pass $L_{1}$ transformer blocks separately for long-form temporal modeling and noise filtering within each modality. 
Each transformer block~\cite{vaswani2017attention} regularly consists of a uni-modal multi-head attention (MHA) and a feed-forward network (FFN) with LayerNorm (LN) and residual connections. 
Afterward, in order to detect modality-aware instances with various lengths, the obtained uni-modal features are fed into an audio-visual pyramid transformer module to integrate informative signals from two modalities at multiple temporal resolutions.
The module consists of $L_{2}$ transformer blocks.
In each block, 2x downsampling using strided convolutions is first applied. Then, the cross-modal multi-head attention (MHA) is conducted by assigning the current modality as the key and value vectors while another as the query vector, which is followed by FFN and LN layers. 
We concatenate the audio-enhanced visual feature $F_{l}^{Va}$ and the visual-enhanced audio feature $F_{l}^{Av}$ from the same pyramid level $l$, and finally get an audio-visual feature pyramid.
\begin{align}
    &Z^{l} = Concat(F_{l}^{Va}, F_{l}^{Av}), \; \\
    &Z = \{Z^{1}, Z^{2}, \dots, Z^{L_{2}}\},
\end{align}
where $l=[1, L_{2}]$, $F_{l}^{Va}, F_{l}^{Av} \in \mathbb{R}^{T_{l}\times D}$, resulting $ Z^{l} \in \mathbb{R}^{T_{l}\times 2D}$, and $T_{l} = T/2^{l-1}$ with a downsampling stride $2^{l-1}$.

\subsection{Task-Specific Experts}
\label{sec: experts}
Unifying these three tasks under a single framework is inherently challenging as they focus on instances with different characteristics. 
Inspired by mixture-of-experts (MoE) networks~\cite{shazeer2017outrageously,bao2022vlmo,mustafa2022multimodal}, we introduce task-specific experts in our transformer blocks.
Unlike previous MoEs that aim to capture modality-specific information (\eg, vision and language), our experts allow the model to learn distinct knowledge for different tasks.
Specifically, as shown in the right of Fig.~\ref{fig:framwork}, the output feature $F_{l-1}, l\in[1, L_{1}+L_{2}]$ from a previous transformer block first passes a shared multi-head attention (MHA) to align information from different tasks. 
Then, the task-specific information can be captured by switching to different task experts.
The task experts compose a Multiway feed-forward network (FFN), where each expert is a standard FFN consisting of two linear layers and an activation.
\begin{align}
& F_{l}^{'}  = \mathrm{MHA}(\mathrm{LN}(F_{l-1}))+ F_{l-1}, \; \\
& F_{l} = \mathrm{Multiway\text{-}FFN}(\mathrm{LN}(F_{l}^{'})) + F_{l}^{'},
\end{align}
where LN is short for layer normalization. $\mathrm{Multiway\text{-}FFN}$ selects the corresponding task experts based on the input data for each task, \eg, if the input video is from the TAL task dataset, we use the TAL expert for data processing.

\subsection{Prediction Head Design}
\label{sec: head}
After the unified audio-visual encoding in Sec.~\ref{sec: encoding}, the output features $\{Z^{l}\}_{l=1}^{L_{2}}$ from the audio-visual feature pyramid will connect to the classification and regression heads to get localization predictions in a single pass.

{\bf Unified language-aware classification head.} 
Since the datasets for different tasks have different label vocabularies, it is a straightforward way to use task-specific heads for classification. 
However, it results in parameter redundancy and limited flexibility due to fixed categories. 
Here, we propose to unify three task-specific classification heads into one by taking advantage of large pre-trained models.
In detail, we treat the instance categories of each dataset as text information and encode them using a pre-trained text encoder.
We highlight that the text encoder is from the same pre-trained tri-modal model~\cite{wang2023one} as the visual and audio encoders used for video tokenization in Sec.~\ref{sec: encoding}. Thus, we can obtain semantically well-aligned embeddings of all three modalities, which provides a strong prior on measuring the similarity between modalities.
To add contextual information, prompts are also customized to help specify labels from different tasks.
The used prompt templates are ``\texttt{A visual event of \{label\}.}'', ``\texttt{An audio visual event of \{label\}.}'', and ``\texttt{A sound event of \{label\}.}'' for TAL, AVEL and SED, respectively. 
The texts are encoded as $\mathcal{T}=\{\mathcal{T}_{i}\}_{i=1}^{N_{k}}$, where $N_{k}$ is the number of classes of the dataset for the $k$-th task.
Then, the encoded texts are linearly projected to a shared embedding space $D'$ with the features from the audio-visual feature pyramid.
We obtain the normalized text vector $\hat{\mathcal{T}}=\{\hat{\mathcal{T}_{i}}\}_{i=1}^{N_{k}} \in \mathbb{R}^{N_{k}\times D'}$ and the normalized cross-modal feature vector $\hat{Z}^{l} \in \mathbb{R}^{T_{l}\times D'}$ from each pyramid level to compute the similarities between them. 

\begin{equation}
   s^{l} = \sigma (\hat{Z}^{l} \hat{\mathcal{T}}^{\top}), 
\end{equation}
where $s^{l}\in \mathbb{R}^{T_{l}\times N_{k}}$ indicates the similarities between $N_{k}$ categories and $T_{l}$ temporal segments in the pyramid level $l$ ($l=[1, L_{2}]$), and $\sigma$ is a learnable scaling factor used to adaptively adjust the magnitude of the similarities as in~\cite{radford2021learning}.
Then, a sigmoid function is attached to predict the probabilities of $N_{k}$ classes at each moment.  
Namely, the class with the largest similarity score is the predicted class of a given segment during inference.

{\bf Regression head.}
We simply apply a lightweight regression head for each task.
Each regression head as in~\cite{geng2023dense} consists of three layers of 1D convolution attached with layer normalization and ReLU activation. 
The parameters of the first two convolutional layers are shared among the three heads. The last layer followed by ReLU outputs the distances to the start/end timestamps of an instance at each moment in the pyramid level $l$ if the instance exists as in Eq.~\ref{equ1}. 

{\bf Loss function.}
Following~\cite{zhang2022actionformer,geng2023dense}, we apply two losses to train our model in an end-to-end manner. 
The sigmoid focal loss~\cite{lin2017focal} $\mathcal{L}_{cls}$ and the generalized IoU loss~\cite{rezatofighi2019generalized} $\mathcal{L}_{reg}$ are used for classification and regression, respectively.
The final loss function for each video is denoted as:
\begin{equation}
    \mathcal{L}=\sum_{t}
    (\frac{1}{T_{all}}\mathcal{L}_{cls} + \frac{\lambda}{T_{+}}\mathbb{I}_{t}\mathcal{L}_{reg}),
\end{equation}
where $T_{all}$ is the total snippet number of all pyramid levels. $\mathbb{I}_{t}$ is an indicator function denoting if a timestamp $t$ contains instances. $T_{+}$ is the number of positive snippets that contain instances across all pyramid levels. $\lambda$ is a coefficient that weights the contribution of $\mathcal{L}_{reg}$. We set $\lambda=1$ by default.

\subsection{Multi-Task Training}
\label{sec: multi-task}
We jointly learn TAL, AVEL and SED tasks in a single, unified architecture by multi-task training. The advantage is that the tasks can potentially learn mutually beneficial knowledge from each other. 
However, the datasets for the tasks vary greatly in size and difficulty, making it very challenging for joint training.
For instance, a single epoch of ActivityNet 1.3~\cite{caba2015activitynet} for TAL task corresponds to around 15 epochs of DESED~\cite{kim2022sound} for SED task. Besides, all videos in DESED are only 10 seconds while the longest video in ActivityNet 1.3 exceeds 12 minutes.
Hence, following the multi-task training method~\cite{lu202012}, we use a Round-Robin Batch-Level Sampling strategy to sample batches from tasks one-by-one in a cyclical manner, where one iteration consists of each task forwarding a batch and updating parameters in sequence.
It effectively trains the model with an equal proportion of individual tasks even though the datasets have various scales (data size and categories), preventing task interference.
The Dynamic Stop-and-Go training scheduler~\cite{lu202012} is also applied to monitor the validation losses of each task to avoid overfitting.
\section{Experiments}
\subsection{Datasets and Evaluation Metrics}
\begin{figure}[!t]
  \centering
  \setlength{\abovecaptionskip}{0.5mm}
   \includegraphics[width=0.7\linewidth]{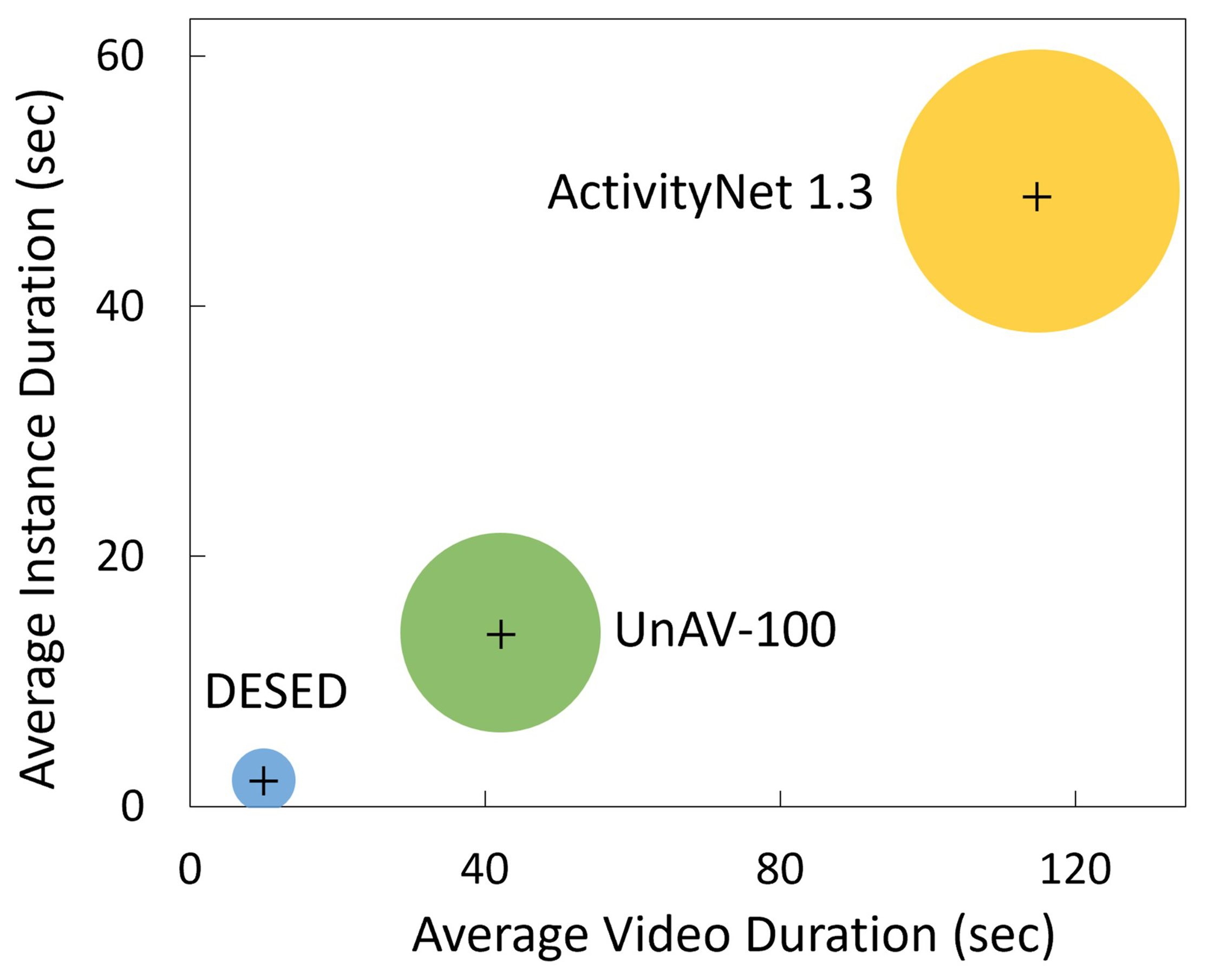}
   \caption{Visualization of dataset statistics. The circle area is proportional to the number of videos in each dataset. There are substantial gaps in dataset scales and video/instance duration between these task-specific datasets.}
   \label{fig:dataset}
\end{figure}

{\bf Temporal action localization.} 
ActivityNet 1.3~\cite{caba2015activitynet} is a large-scale benchmark for video action localization. It contains 200 common human activity classes and around 20k untrimmed videos collected from YouTube.
Following the common practice~\cite{lin2021learning,liu2022end,zhang2022actionformer}, we use its training set during training and report the performance on the validation set. 
The standard evaluation metric for TAL is mean Average Precision (mAP). We use the mAPs at the tIoU thresholds [0.5:0.05:0.95] and the average mAP is also reported. 

{\bf Sound event detection.}
DESED~\cite{turpault:hal-02160855} is the benchmark for the DCASE Challenge~\cite{mesaros2017dcase}. 
It consists of 10-second audio clips with 10 sound event classes in domestic environments.
DESED just has sound data and provides strong annotations (\ie, classes and temporal boundaries) for its recorded validation set (1,168 clips) and public evaluation set (692 clips).
We downloaded their original videos with audio from YouTube.
We use the validation set for training and report the results on the public evaluation set.
Since the traditional evaluation metric~\cite{mesaros2016metrics} is only suitable for the sound-only clip-level task with extremely fine granularity (\ie, 0.05s per clip) on audio spectrograms, we use mAPs@[0.5:0.2:0.9] and report the average mAP of mAPs@[0.1:0.1:0.9] in our experiments.

{\bf Audio-visual event localization.}
UnAV-100~\cite{geng2023dense} consists of 10,790 untrimmed videos with around 30k audio-visual events.
It covers 100 event categories spanning a wide range of domains, \eg, human activities, animal/natural sounds, \etc. 
Following~\cite{geng2023dense}, we use its train split for training and test split for testing. 
For evaluation, we use the mAPs at the tIoU thresholds [0.5:0.1:0.9] and also report the average mAP between 0.1 and 0.9 with the step of 0.1 (\ie [0.1:0.1:0.9]).

There exist significant gaps between these three task-specific datasets in video/instance duration and data scales as shown in Fig.~\ref{fig:dataset}, which makes the unification of TAL, AVEL and SED tasks a very challenging problem.
Specifically, ActivityNet 1.3 for TAL has many more videos with an average event duration of about 50 seconds, while DESED for SED has fewer videos with events averaging only 1 second. Training on such data naturally leads the TAL task to benefit more from long-duration events, while the SED task relies on short-term signals. 
We chose these datasets as they are currently the most mainstream datasets with relatively high quality. 
ActivityNet 1.3 has rich common human activities, 
which has domain overlaps with UnAV-100 and DESED, facilitating the model to learn mutually beneficial knowledge across tasks during multi-task training. 
Besides, only UnAV-100 for AVEL is based on untrimmed videos, and DESED is the only dataset whose audio and visual modalities are both available for SED task.

\begin{table*}
\begin{center}
 \caption{Comparison of our multi-task models to single-task performance. 
 We use both audio and visual modalities for all models. ``ST'': single-task, ``AT'': all tasks.
 ``All Tasks Average'' is computed by averaging the average mAP results of all three tasks. ``$\cdot \mathrm{M} (\cdot)$'': total $\cdot \mathrm{M}$ parameters needed for $\cdot$ models to conduct all three tasks.}
 \vspace{-2.5mm}
  \resizebox{\linewidth}{!}{
  \begin{tabular}{ll|cccc|cccccc|cccc|c|r}
    \toprule
    && \multicolumn{4}{c}{ActivityNet 1.3 (TAL)} & \multicolumn{6}{c}{UnAV-100 (AVEL)} & \multicolumn{4}{c}{DESED (SED)} &\multirow{2}{*}{\thead{All Tasks \\ Average}}  \\
    \cmidrule(lr){3-6} \cmidrule(lr){7-12}\cmidrule(lr){13-16}
     && 0.5 & 0.75 & 0.95 & Avg. & 0.5 & 0.6 & 0.7 & 0.8 & 0.9 & Avg. & 0.5 & 0.7 & 0.9 & Avg. &  &\# params  \\
    \midrule
    {\color{gray}1}&Single-Task (ST)&56.6&35.4&5.1&35.3&53.2&46.7&39.9&31.6&19.8&49.6&61.0&45.9&19.8&57.7& 47.5& 186M (3) \\
    {\color{gray}2}&Naive Multi-Task (Base) &55.4&34.9&6.3&34.8&53.0&47.4&41.0&33.5&21.5&49.8&62.7&50.5&26.2&58.6& 47.7 &62M (1) \\
    \midrule
    {\color{gray}3}&Ours (TAL \& AVEL) &56.4&35.7&5.0&35.7&54.0&48.9&41.9&34.0&21.7&50.8&-&-&-&-&-&97M (1) \\
    {\color{gray}4}&Ours (TAL \& SED) &56.5&36.1&4.2&35.5&-&-&-&-&-&-&61.8&49.3&26.1&58.2&-&97M (1)\\
    {\color{gray}5}&Ours (AVEL \& SED) &-&-&-&-&51.8&46.8&40.1&30.0&18.2&48.3&62.6&49.4&{\bf27.2}&59.4&-&97M (1) \\
    {\color{gray}6}&Ours$_{AT}$ (All-Task) & {\bf 57.1}&{\bf36.4}&5.5&36.1&54.1&48.6&42.1&34.3&20.5&50.7&64.0&49.5&27.0&59.8&48.9 &130M (1) \\
    \midrule
    {\color{gray}7}&
    Ours$_{AT\rightarrow ST}$ 
    &56.8&36.0&{\bf 6.7}&{\bf36.2}&{\bf54.8}&{\bf49.4}&{\bf43.2}&{\bf35.3}&{\bf22.5}&{\bf51.7}&{\bf65.1}&{\bf50.9}&26.1&{\bf61.1}&{\bf49.7} &186M (3)\\
    \bottomrule
  \end{tabular}}
  \label{tab:multitask}
  \vspace{-4mm}
  \end{center}
\end{table*}

\subsection{Implementation Details}
\label{sec: details}
The sampling rates of sounds and video frames are 16 kHz and 16 fps, respectively. 
We feed 16 consecutive frames using a sliding window with stride 8 for ActivityNet 1.3, and downsample the features into a fixed length of 256 following~\cite{zhang2022actionformer}. For UnAV-100 and DESED, we use the stride of 4, and pad or crop the feature sequences to 256 and 64, respectively. For each corresponding 1s audio segment, the same stride duration (0.5/0.25s) is used to temporally align with the visual ones. 
The visual, audio and text encoders of ONE-PEACE~\cite{wang2023one} are utilized to extract semantically aligned three modality embeddings.
The visual encoder is further fine-tuned on Kinetics-400~\cite{carreira2017quo} by freezing the pre-trained parameters and adding several MLP layers, enhancing temporal awareness while maintaining good alignment with other modalities.
The extracted feature dimension is 1,536 for all three modalities. 
In the audio-visual pyramid transformer, the number of attention heads is 4 in both uni-modal and cross-modal blocks. The temporal downsampling operation is realized by using a single depth-wise 1D convolution. 
The dimension of the embedding space in the framework is $D=D^{'}=512$, and $L_{1}=2$, $L_{2}=6$. 
During training, we use Adam for optimization and simply set the same hyperparameters for all datasets of the three tasks. 
Specifically, the mini-batch size is 16, the initial learning rate is 1e-3 and a cosine learning rate decay is used.
Our model is trained only for 5 epochs with a linear warmup of 2 epochs and the weight decay is 1e-4. 
During inference, our model outputs the on/offsets and classes with confidence scores for all three types of instances occurring in a given video. The output candidates are then processed by Soft-NMS~\cite{bodla2017soft} to eliminate highly overlapping ones.
\begin{table}
\centering
 \caption{Cross-task transfer performances. Only the head layers of the source-task models are fine-tuned for target tasks during transfer. The results of the single-task models without transfer are in bold.}
  \resizebox{1.0\linewidth}{!}{
  \begin{tabular}{c|cccccc}   
    \toprule
    \multirow{2}{*}{\diagbox{Source Task}{Target Task}}
    & \multicolumn{2}{c}{TAL} & \multicolumn{2}{c}{AVEL} & \multicolumn{2}{c}{SED} \\
    \cmidrule(lr){2-3} \cmidrule(lr){4-5}\cmidrule(lr){6-7}
    & 0.5 & Avg. & 0.5 & Avg. & 0.5 & Avg. \\
    \midrule
    TAL & {\bf56.6} & {\bf35.3} & 30.6 & 29.8  &28.1 & 27.8  \\
    AVEL & 35.2 &21.3  & {\bf53.3} & {\bf49.6} & 43.4  & 42.9  \\
    SED & 2.5 &1.3  & 7.4 &9.0 & {\bf61.0} & {\bf57.7} \\
    \bottomrule
  \end{tabular}}
  \label{tab:transfer}
  \vspace{-4mm}
\end{table}

\subsection{Results and Analysis}
{\bf Single-task v.s. multi-task.}
To demonstrate the effectiveness of our unified framework, in Table~\ref{tab:multitask}, we compare our all-task (AT) model (row 6) with two baselines, \ie, single-task (ST) and naive multi-task (Base) in row 1-2. 
Single-task (ST) models were trained individually on the three tasks, using standard transformer blocks with one FFN and a 1D-conv classifier similar to the regressor.
The naive multi-task (Base) model has the same architecture as ST models except for using multi-task training strategies in Section~\ref{sec: multi-task}. 
In other words, the Base model does not use proposed task-specific experts and the language-aware classifier compared with our AT model.
We use the same hyperparameters described in Section~\ref{sec: details} as the default setting unless specified otherwise.
We observe that naive multi-tasking cannot get decent results on the three tasks (row 2). 
The performance of the Base model on SED and AVEL saw a slight improvement, due to beneficial knowledge learned from each other. However, the results on TAL declined since there exists a significant gap between ActivityNet 1.3 and other datasets.   
In contrast, our all-task (AT) model (row 6) that applies the task-specific experts and the unified language-aware classifier can achieve performance boosts on all three tasks compared with ST models, \eg, $+0.8\%$, $+1.1\%$ and $+2.1\%$ at the average mAP on TAL, AVEL and SED, respectively, and $+1.4\%$ at the average performance of all tasks.
Besides, the total number of parameters reduces by a factor of $1.4 \times$ (\ie, 186M to 130M), going from 3 full models to only 1 required for all tasks. 
It implies both the effectiveness and efficiency of our model.

\begin{table*}[!t]
  \centering
 \caption{Comparison with existing state-of-the-art methods. We report the mAP at tIoU=0.5 and the average mAP on three tasks. Best results are in bold and second best \underline{underlined}. ``OP-V/A'' denotes the visual/audio encoder of ONE-PEACE~\cite{wang2023one}. ``*'' denotes that the results  
  for the AVEL task are from UnAV~\cite{geng2023dense}.
  }
  \resizebox{0.9\linewidth}{!}{
  \begin{tabular}{lccccccccc}
    \toprule
    \multirow{2.2}{*}{Method} & \multirow{2.2}{*}{Visual Encoder} &\multirow{2.2}{*}{Audio Encoder}& \multicolumn{2}{c}{ActivityNet 1.3 (TAL)} & \multicolumn{2}{c}{UnAV-100 (AVEL)} & \multicolumn{2}{c}{DESED (SED)} &\multirow{2.2}{*}{\thead{All Tasks \\ Average}} \\
    \cmidrule(lr){4-5} \cmidrule(lr){6-7}\cmidrule(lr){8-9}
     &&& 0.5 & Avg. & 0.5 & Avg. & 0.5 & Avg.  \\
    \midrule
    SSN~\cite{zhao2017temporal}& I3D~\cite{carreira2017quo} &-& 39.1&24.0&-&-& - & -&- \\
    TAL-Net~\cite{chao2018rethinking}&I3D~\cite{carreira2017quo}&-& 38.2&20.2&-&-& - & -&- \\
    P-GCN~\cite{zeng2019graph} & I3D~\cite{carreira2017quo} &-& 42.9& 27.0&-&-& - & -&- \\
    PCG-TAL~\cite{su2020pcg} & I3D~\cite{carreira2017quo} & -&42.1& 27.3&-&-& - & -&- \\
    SSN-GCM~\cite{zeng2021graph} & I3D~\cite{carreira2017quo} & - & 42.6 & 27.2 &-&-& - & -&-  \\
    TadTR~\cite{liu2022end} & I3D~\cite{carreira2017quo} & -&43.7 & 29.9&-&-& - & -&- \\
    ActionFormer~\cite{zhang2022actionformer}& I3D~\cite{carreira2017quo} &-&46.1&30.5&-&-& - & -&-\\
    TriDet~\cite{shi2023tridet} & I3D~\cite{carreira2017quo} &-& 48.5 & 31.1&-&-& - & -&- \\
    TE-TAD~\cite{kim2024te} & I3D~\cite{carreira2017quo} & - & 48.0 & 31.4 &-&-& - & -&- \\
    ActionFormer~\cite{zhang2022actionformer}& - &VGGish~\cite{hershey2017cnn}&-&-&-&-&39.6&37.8&-\\
    \midrule
    VSGN~\cite{zhao2021video}*&I3D~\cite{carreira2017quo} & VGGish~\cite{hershey2017cnn}&-&-&24.5&24.1 & - & -&-\\
    TadTR~\cite{liu2022end}*& I3D~\cite{carreira2017quo} &VGGish~\cite{hershey2017cnn}&- &-&30.4&29.4& - & -&-\\
    ActionFormer~\cite{zhang2022actionformer}& I3D~\cite{carreira2017quo}& VGGish~\cite{hershey2017cnn}&47.2 &31.1 & 43.5&42.2&42.2&39.7 & 37.7 \\
    TriDet~\cite{shi2023tridet} & I3D~\cite{carreira2017quo} &VGGish~\cite{hershey2017cnn}& 49.3 &32.1 & 46.2&44.4&42.0&41.2 & 39.2 \\
    TE-TAD~\cite{kim2024te} & I3D~\cite{carreira2017quo} & VGGish~\cite{hershey2017cnn} & 49.1 & 32.5 &45.8&44.8& 41.9 & 42.5&39.9 \\
    UnAV~\cite{geng2023dense}& I3D~\cite{carreira2017quo} &VGGish~\cite{hershey2017cnn} &42.7&28.1&50.6&47.8&51.6&48.8 & 41.6 \\
    \midrule
    ActionFormer~\cite{zhang2022actionformer}& OP-V~\cite{wang2023one} &OP-A~\cite{wang2023one}&55.2&35.4&49.2 &47.0& 48.2& 44.6 & 42.3\\
    TriDet~\cite{shi2023tridet} & OP-V~\cite{wang2023one} &OP-A~\cite{wang2023one}& \underline{56.9} & 35.9 & 49.7&47.3& 48.3&46.0& 43.1 \\
    TE-TAD~\cite{kim2024te} & OP-V~\cite{wang2023one} &OP-A~\cite{wang2023one} & 56.1 & 36.0 & 49.0 & 47.8 & 48.0 & 46.5 & 43.4  \\
    UnAV~\cite{geng2023dense}& OP-V~\cite{wang2023one} &OP-A~\cite{wang2023one} &50.5&32.5&53.8 & \underline{51.0}&60.9&57.8& 47.1\\
    Ours$_{AT}$ & OP-V~\cite{wang2023one} &OP-A~\cite{wang2023one}&{\bf 57.1}&\underline{36.1}&\underline{54.1}&50.7&\underline{64.0}& \underline{59.8} & \underline{48.9}\\
    Ours$_{AT\rightarrow ST}$ & OP-V~\cite{wang2023one}&OP-A~\cite{wang2023one} & 56.8&{\bf36.2}&{\bf 54.8}&{\bf51.7}&{\bf65.1} & {\bf61.1} & {\bf49.7}\\
    \bottomrule
  \end{tabular}}
  \label{tab:sota_new}
\end{table*}

{\bf Cross-task transfer.} We perform transfer learning across tasks. We only fine-tune the last layer of the classification and regression heads of the single-task (ST) models during transfer. In Table~\ref{tab:transfer}, we can see that transfer learning yields very poor results, especially, when transferring the SED single-task model for TAL and AVEL tasks. 
Besides, cross-task transfer between TAL and AVEL also shows significant performance drops compared to their single-task models, \eg, $-19.8\%$ at the average mAP when transferring TAL to AVEL compared to the results of the AVEL ST model. 
It demonstrates the substantial task/data gap between tasks, thus the unification of the three tasks is a challenging and non-trivial problem.

{\bf Pair-wise task relationships.}
We also explore pair-wise task relationships by jointly training two of three tasks, as shown in rows 3-5 in Table~\ref{tab:multitask}.
We observe that when applying our proposed task-specific experts and unified language-aware classifier, both AVEL and SED (row 3-4) can benefit from the rich common instances in ActivityNet 1.3 when jointly trained with TAL, leading to an obvious improvement compared with the single-task models, \ie, $+1.2\%$ and $+0.5\%$ at the average mAP, respectively.
Besides, SED gains a significant boost when trained with AVEL (row 5), \ie, $+1.7\%$ at the average mAP.
It could be attributed to category overlap between UnAV-100 and DESED, and training with the large dataset can help prevent overfitting in the small one.  
Conversely, due to the large gap in instance duration and dataset scales, SED tends to have a negative effect on the AVEL task, resulting in a decrease of $1.3\%$ at the average mAP (row 5). But this effect can be regulated by jointly training all three tasks together using our proposed AT model (row 6). 

{\bf Multi-task learning as pre-training.}
Inspired by~\cite{lu202012}, we fine-tune each single-task model on our trained AT model to demonstrate that the AT model can allow downstream tasks to take advantage of multi-task training. 
The results are shown in row 7 of Table~\ref{tab:multitask}, where we initialize ST models using the trained AT model and fine-tune them individually using the same training recipe as in row 1.
We can see that the ST models fine-tuned on our AT model outperform the single-task models in row 1 by a large margin, \ie, $+0.9\%$, $+2.1\%$ and $+3.4\%$ on TAL, AVEL and SED, respectively. 
It indicates that joint training can capture knowledge that is mutually beneficial to all these three tasks, being an effective pre-training step for single-task models.

\begin{table*}[t]
  \centering
\caption{Ablation study on the main proposed components.
  ``E-$L_{1}$'' denotes applying task-specific experts on the early $L_{1}$ transformer blocks, and ``E-$L_{2}$'' denotes applying experts on the later $L_{2}$ blocks.
  ``LCH'' is short for language-aware classification head. ``Prompt'' denotes using prompts when tokenizing instance categories for different tasks. }
  \resizebox{1.0\linewidth}{!}{
  \begin{tabular}{ccc|cc|cccc|cccccc|cccc|c}
    \toprule
    &&&&&\multicolumn{4}{c}{TAL}&\multicolumn{6}{c}{AVEL}&\multicolumn{4}{c}{SED
    } & \\
    \cmidrule(lr){6-9}\cmidrule(lr){10-15} \cmidrule(lr){16-19} 
    &E-$L_{1}$ & E-$L_{2}$ & LCH & Prompt & 0.5 & 0.75 & 0.95 & Avg. & 0.5 & 0.6 & 0.7 & 0.8 & 0.9 & Avg. & 0.5 & 0.7 & 0.9 & Avg.&\#params \\
    \midrule
    {\color{gray}1}&&&&& 55.4 & 34.9 &{\bf6.3} & 34.8 & 53.0 & 47.4 & 41.0 & 33.5 & 21.5 & 49.8 & 62.7 & {\bf50.5} & 26.2 & 58.6 & 62M \\
    \midrule
{\color{gray}2}&&&\checkmark&\checkmark& 55.8 & 35.2 & 5.2 & 35.0 & 52.9 & 47.7 & 41.6 & 33.5 & 21.3 & 49.9 & 63.5 & 49.8 & {\bf27.3} & 59.7 & 64M\\
    {\color{gray}3}&\checkmark& &\checkmark&\checkmark&56.9& 36.1 & 5.7 & 35.9 &53.7 & 48.5 & 42.5 & 33.9 & 20.5 & 50.5 & 61.4 & 49.8 & 26.4 & 58.5 & 80M\\
    {\color{gray}4}&& \checkmark& \checkmark&\checkmark&56.8& 36.1 & 6.2 & 35.9 & 54.5 & 49.2 & 42.5 & 34.6 & 20.8 & 50.8 & 62.9 & 49.3 & 27.2 & 59.3 & 114M \\
    \midrule
    {\color{gray}5}&\checkmark & \checkmark & & &56.2& 35.7 & 5.1 &35.2&{\bf54.6}& 49.4&42.4&34.6&{\bf22.6}&{\bf51.2}&57.4&43.5&19.5&54.8&133M\\
    {\color{gray}6}&\checkmark & \checkmark & \checkmark& &57.0&36.3&5.2&35.9&54.2&{\bf49.6}&{\bf42.8}&{\bf34.8}&21.3&51.0&63.7&49.8&26.6&59.7&130M\\
    \midrule
    {\color{gray}7}&\checkmark & \checkmark & \checkmark&\checkmark&{\bf57.1}&{\bf36.4}& 5.5 & {\bf36.1} &54.1&48.6& 42.1&34.3&20.5& 50.7&{\bf64.0}&49.5&27.0&{\bf59.8}&130M\\
    \bottomrule 
  \end{tabular}}
  \label{tab:expert-head}
  \vspace{-2mm}
\end{table*}

\subsection{Comparison with Existing Methods}
Table~\ref{tab:sota_new} presents the comparison results of our model with state-of-the-art works.
We highlight that our UniAV (Ours$_{AT}$) is the only model that performs all three tasks by a single model parameters, while all other models in Table~\ref{tab:sota_new} are task-specific models that are separately trained for each task.
For TAL task, we note that many previous TAL methods achieved superior results on ActivityNet 1.3 by combining with the external action classifier~\cite{xiong2016cuhk,wang2017untrimmednets}. 
By contrast, our model has good capabilities in both classification and regression and does not need to rely on any external classification models. 
For a fair comparison, we list the results of methods without using external classifiers in Table~\ref{tab:sota_new}.
We simply concatenate audio and visual features as input for those TAL methods~\cite{liu2022end,shi2023tridet,zhang2022actionformer,zhao2021video, kim2024te} when using both modalities.
We can see that, with ONE-PEACE~\cite{wang2023one} features, our all-task (AT) model reaches an average mAP of $36.1\%$, outperforming other state-of-the-art task-specific models.
We also trained UnAV~\cite{geng2023dense} that is tailored for AVEL task to conduct TAL task, but found the result is much lower than highly specialized TAL models~\cite{shi2023tridet,zhang2022actionformer}. It indicates that the task-specific models cannot be generalized effectively to other tasks, while our unified model can achieve superior or on-par performances on all three tasks with good generalizability.
For AVEL task, our AT model gets very competitive results compared to UnAV~\cite{geng2023dense} when using the same ONE-PEACE features. 
Besides, our single-task fine-tuned model (Ours$_{AT\rightarrow ST}$) further improves the average mAP to $51.7\%$, setting a new state-of-the-art result on the AVEL task.
For SED task, since the traditional SED methods only support super fine-grained sound spectrograms as input with the evaluation metric not suitable for our unified approach, we implemented Actionformer~\cite{zhang2022actionformer}, TriDet~\cite{shi2023tridet} and UnAV~\cite{geng2023dense} to conduct SED task. We can see that our AT model outperforms all other methods by a large margin. 
Overall, we emphasize that our proposed AT model (UniAV) holds superior or on-par performances on all three tasks using only a single unified model, and achieves the best All Tasks Average score compared with other methods, which significantly distinguishes our model from other task-specific models. 


\subsection{Ablation Studies}
{\bf Effect of task-specific experts (TE) and unified language-aware classification head (LCH).}
First, we explore the effect of our TE on different layers of the pyramid transformer.
As shown in Table~\ref{tab:expert-head}, applying TE in the later $L_{2}$ cross-modal transformer blocks (row 4) has better results than that in the early $L_{1}$ uni-modal transformer blocks (row 3). It indicates that the later stages of cross-modal fusion can capture distinct knowledge that is more beneficial for each task. 
Besides, adding experts in all transformer blocks (row 7)  can achieve the best performances on TAL and AVEL tasks.
Second, we find that using separate task-specific classifiers (row 5) leads to quite unstable results. 
In contrast, notable improvements are observed on TAL and SED when using LCH with only categories as text tokens (row 6), especially boosting SED with a $4.5\%$ increase at average mAP. 
It could be attributed to the LCH based on the large language encoder~\cite{wang2023one}, which enhances generalization and effectively avoids overfitting on the small dataset.
Overall, combining both modules yields complementary effects, significantly improving all tasks (row 1 vs. row 7), with greater performance gains at lower thresholds. 
\vspace{-4mm}

\begin{table}
  \centering
 \caption{Ablation study on the effect of parameter quantity. ``TE'' is short for task-specific experts. ``$D \& D'$'' denotes the dimension of the transformer blocks.}
  \resizebox{1.0\linewidth}{!}{
  \begin{tabular}{c|c|cccccc|c}
    \toprule
    \multirow{3}{*}{TE}&\multirow{3}{*}{$D \& D'$} &\multicolumn{2}{c}{TAL}&\multicolumn{2}{c}{AVEL}&\multicolumn{2}{c}{SED
    } \\
    \cmidrule(lr){3-4} \cmidrule(lr){5-6}\cmidrule(lr){7-8}
     && 0.5 & Avg. & 0.5 & Avg. & 0.5 & Avg. & \# params \\
    \midrule
    & 512 & 55.8 & 35.0 & 52.9 & 49.9 & 63.5 & 59.7 & 64M\\
    & 640 & 55.0 & 34.5 & 52.3 & 49.4 & 63.4 & 59.6 & 95M  \\
    & 768 & 55.1 & 34.9 & 52.5 & 49.4 & 62.4 & 59.3 & 132M \\
    \midrule
    \checkmark&512&{\bf57.1}&{\bf36.1}&{\bf54.1}&{\bf50.7}&{\bf64.0}&{\bf59.8}&130M \\
    \bottomrule 
  \end{tabular}}
  \label{tab:param}
\end{table}

\begin{table}[t]
  \centering
    \caption{Ablation study on our models using I3D~\cite{carreira2017quo} and VGGish~\cite{hershey2017cnn} features. ``SOTA'': TriDet~\cite{shi2023tridet} for TAL, UnAV~\cite{geng2023dense} for AVEL and SED using same features.}
  \resizebox{1.0\linewidth}{!}{
  \begin{tabular}{l|cccccc}
    \toprule
    \multirow{3}{*}{Method}& \multicolumn{2}{c}{TAL} & \multicolumn{2}{c}{AVEL} & \multicolumn{2}{c}{SED} \\
    \cmidrule(lr){2-3} \cmidrule(lr){4-5}\cmidrule(lr){6-7}
    & 0.5 & Avg. & 0.5 & Avg. & 0.5 & Avg. \\
    \midrule
    SOTA & {\bf 49.3} & 32.1 & {\bf 50.6} & 47.8 & 51.6 & 48.8 \\
    Ours$_{ST}$ & 48.8 & 31.8 & 48.8 & 46.7 & 50.5 & 48.7 \\
    Ours$_{AT}$ & 48.4 & 31.9 & 49.3 & 47.0 & 49.8 & 49.2 \\
    Ours$_{AT_{w/o LCH}}$ & 48.9 & 32.4 & 49.6 & 47.7 & 50.9 & 50.0 \\
    Ours$_{AT_{w/o LCH} \rightarrow ST}$ & 49.0 & {\bf 32.6} & 50.1 & {\bf 48.2} & {\bf 52.4} & {\bf 50.6} \\
    \bottomrule 
  \end{tabular}}
  \label{tab:i3d-vggish}
    \vspace{-3mm}
\end{table}

\begin{table}[t]
\centering
  \caption{Ablation study on our models using ImageBind~\cite{girdhar2023imagebind} and LanguageBind~\cite{zhu2023languagebind} as encoders. ``TE" and ``LCH" denote the task-specific experts and the language-aware classification head, respectively.}
  \resizebox{1.0\linewidth}{!}{ 
  \begin{tabular}{c|c|cc|cc|cc}
    \toprule
    \multirow{2}{*}{Encoder}& \multirow{2}{*}{Method}&  \multicolumn{2}{c}{TAL} & \multicolumn{2}{c}{AVEL} & \multicolumn{2}{c}{SED} \\
    \cmidrule(lr){3-4} \cmidrule(lr){5-6}\cmidrule(lr){7-8}
    & & 0.5 & Avg. & 0.5 & Avg. & 0.5 & Avg. \\
    \midrule
    \multirow{5}{*}{ImageBind~\cite{girdhar2023imagebind}} & ST& 49.3 & 30.8 & 46.4 & 43.6 & 55.4 & 51.8 \\
    & Base & 48.4 & 30.2 & 45.9 & 43.7 & 56.0 & 52.2 \\
    &w/ TE  & 49.4 & 31.3 & {\bf48.8} & {\bf46.0} & 56.5  & 54.0 \\
    &w/ LCH & 49.1 & 31.2 & 46.8 & 44.6 & 56.6 & 53.4 \\
    & Ours$_{AT}$ &  {\bf50.6} & {\bf31.8} & 48.7 & 45.8 & {\bf57.1} & {\bf54.1} \\
   \midrule
    \multirow{5}{*}{LanguageBind~\cite{zhu2023languagebind}} &ST & 50.6 & 31.4 & 48.2 & 44.8 & 57.0 & 54.0 \\
    & Base & 49.8 & 31.0 & 47.5 & 44.2 & 57.8 & 54.6 \\
    & w/ TE &51.3 & 32.4 & {\bf 50.4} & 47.3  & 59.5 & 56.8 \\
    &w/ LCH &50.9 & 32.1 & 49.0 & 46.1 & 59.6 & 56.8 \\
    & Ours$_{AT}$ & {\bf 53.0} & {\bf 33.5} & 50.3 & {\bf47.5} & {\bf 60.1} & {\bf 57.3} \\
    \bottomrule 
  \end{tabular}}
  \label{tab:bind}
\end{table}

\begin{table}[t]
    \centering
     \caption{Ablation study on the effect of audio-visual fusion for TAL and SED tasks.
  ``V'' denotes visual-only, ``A'' denotes audio-only, and ``A\&V''denotes both audio and visual.}
    \resizebox{0.7\linewidth}{!}
    {
  \begin{tabular}{c|c|cccc}
    \toprule
    \multirow{4}{*}{TAL}&Modality&0.5&0.75&0.95&Avg.\\
    \cmidrule(lr){2-6}
     & V &55.0&35.0&4.8&34.2  \\
    & A\&V & {\bf56.6}&{\bf35.4}&{\bf 5.1}&{\bf 35.3}\\
    \midrule
    \multirow{4}{*}{SED}& Modality& 0.5&0.7&0.9&Avg. \\
    \cmidrule(lr){2-6}
     & A & 51.4&38.1&13.9&49.5\\
    & A\&V & {\bf61.0}&{\bf45.9}&{\bf19.8}&{\bf57.7}\\
    \bottomrule 
  \end{tabular}}
  \label{subtab:fusion}
  \vspace{-3mm}
\end{table}

\begin{table}[t]
    \centering
    \caption{Ablation study on the importance of motion information during visual tokenization.
  ``FT'' denotes using the visual encoder of ONE-PEACE fine-tuned on Kinetics-400.}
    {
  \begin{tabular}{cc|cccccc}
    \toprule
    &&\multicolumn{2}{c}{TAL}&\multicolumn{2}{c}{AVEL}&\multicolumn{2}{c}{SED
    } \\
    \cmidrule(lr){3-4} \cmidrule(lr){5-6}\cmidrule(lr){7-8}
    Model & FT & 0.5 & Avg. & 0.5 & Avg. & 0.5 & Avg. \\
    \midrule
    ST & &50.5& 31.9 & 51.4&48.4&58.7&55.9\\
    ST & \checkmark &{\bf56.6} & {\bf35.3} &{\bf53.2}&{\bf49.6}&{\bf61.0}&{\bf57.7}\\
    \midrule
    AT &  &51.7&32.6&50.7& 48.6 &61.8&59.3 \\
    AT & \checkmark & {\bf57.1}&{\bf36.1}&{\bf54.1}&{\bf50.7}&{\bf64.0}&{\bf59.8}\\
    \bottomrule 
  \end{tabular}}
  \label{subtab:kinetics}
  \vspace{-2mm}
\end{table}
\label{tab:ablation study}

\begin{table}[t]
  \centering
  \caption{Ablation study on other large-scale audio and visual encoders. We show the results on the public evaluation set of DESED dataset for SED task.}
  \resizebox{1.0\linewidth}{!}{
  \begin{tabular}{c|c|cccc}
    \toprule
    Modality&Encoder&0.5&0.7&0.9&Avg.\\
    \midrule
    \multirow{4}{*}{A} & CLAP~\cite{wu2023large} &27.2&16.5&5.1&26.2  \\
    & \textcolor{black}{ImageBind~\cite{girdhar2023imagebind}} & \textcolor{black}{39.3} & \textcolor{black}{28.6} & \textcolor{black}{12.4} & \textcolor{black}{37.1} \\
    & \textcolor{black}{LanguageBind~\cite{zhu2023languagebind}} & \textcolor{black}{44.5} & \textcolor{black}{34.4} & \textcolor{black}{13.5} & \textcolor{black}{42.3} \\ 
    & ONE-PEACE~\cite{wang2023one} &51.4&38.1&13.9&49.5 \\
    \midrule
    \multirow{4}{*}{V} & CLIP~\cite{radford2021learning} & 29.2&18.6&8.9&30.6\\
    & \textcolor{black}{ImageBind~\cite{girdhar2023imagebind}} & \textcolor{black}{32.3} & \textcolor{black}{20.3} & \textcolor{black}{9.5} & \textcolor{black}{32.5} \\
    & \textcolor{black}{LanguageBind~\cite{zhu2023languagebind}} & \textcolor{black}{30.6} & \textcolor{black}{18.6} & \textcolor{black}{8.5} & \textcolor{black}{31.2} \\ 
    & ONE-PEACE~\cite{wang2023one} &29.9&15.2&4.8&29.8 \\
    \midrule
    \multirow{4}{*}{A\&V} & CLAP~\cite{wu2023large} \& CLIP~\cite{radford2021learning}&52.6&39.2&{\bf20.0}&50.1\\
    & \textcolor{black}{ImageBind~\cite{girdhar2023imagebind}} & \textcolor{black}{55.4} & \textcolor{black}{41.1} & \textcolor{black}{18.3} & \textcolor{black}{51.8} \\
    & \textcolor{black}{LanguageBind~\cite{zhu2023languagebind}} & \textcolor{black}{57.0} & \textcolor{black}{44.5} & \textcolor{black}{19.4} & \textcolor{black}{54.0} \\ 
    & ONE-PEACE~\cite{wang2023one} &{\bf61.0}&{\bf45.9}&19.8&{\bf57.7}\\
    \bottomrule 
  \end{tabular}}
  \label{tab:encoders}
  \vspace{-3mm}
\end{table}

{\bf Effect of parameter quantity.}
From the study of the proposed task-specific experts (TE) in Table~\ref{tab:expert-head}, we can see that adding experts on more transformer blocks leads to a significant increase in the number of parameters. 
In Table~\ref{tab:param}, we explore the effect of parameter quantity on our model. We can observe that when removing the task-specific experts and increasing the dimensions $D$ and $D'$, the model’s performance does not improve but declines across the three tasks.
It clearly proves that the performance improvement of our model is not due to an increase in parameters but the effectiveness of our proposed task-specific experts.

{\bf Effect of I3D and VGGish features.}
We also evaluate our model using I3D~\cite{carreira2017quo} and VGGish~\cite{hershey2017cnn} features that are commonly used on previous models.
The results are shown in Table~\ref{tab:i3d-vggish}. 
Note that the I3D visual and VGGish audio embeddings are not semantically aligned with the used ONE-PEACE~\cite{wang2023one} text embeddings, which severely limits the effectiveness of our unified language-aware classifier (LCH). 
Alternatively, our AT model without LCH achieves performance boosts on all three tasks compared to our single-task (ST) models, indicating the great benefit of the proposed task-specific experts. 
Besides, fine-tuning gains further performance boosts, setting new state-of-the-art results compared with existing methods with the same features.
In conclusion, the lack of modality semantic alignment (audio-visual-language) and limited generalization of the traditional I3D and VGGish encoders constrain our model's capability and flexibility.
It clearly proves the great necessity of using a general model pre-trained by aligning vision, audio and language modalities (\eg, ONE-PEACE) as our audio/visual encoders.

{\bf Effect of other tri-modal aligned encoders.}
We also demonstrate that our model remains effective when using other existing tri-modal aligned encoders, including ImageBind~\cite{girdhar2023imagebind} and LanguageBind~\cite{zhu2023languagebind}. We apply the same sampling strategy used for ONE-PEACE to extract features for a fair comparison.
As shown in Table~\ref{tab:bind}, our AT model consistently outperforms the Naive Multi-Task (Base) model across all three tasks when using both encoders. Besides, we can see that the proposed task-specific experts (TE) and the language-aware classification head (LCH) maintain their remarkable effectiveness. Therefore, our model does not rely on the specific encoder~\cite{wang2023one}, rather, the semantically well-aligned embeddings are what support the effectiveness and flexibility of our model.
We note that since the architecture and training strategy of ONE-PEACE~\cite{wang2023one} uniquely emphasize modality interaction and fine-grained understanding, along with the visual encoder fine-tuned on Kinetics-400~\cite{carreira2017quo}, it leads to superior encoding performance compared to the other tri-modal encoders.

{\bf Audio-visual fusion for TAL and SED.}
We also verify the effectiveness of audio-visual fusion for TAL and SED tasks in Table~\ref{subtab:fusion}. We use the single-task ST model (row 1 in Table~\ref{tab:multitask}) for each task to conduct the experiments. 
For TAL, we can see the performance increase ($+1.1\%$ at average mAP) as we insert audio signals to apply cross-modal interactions.
For SED, the model obtains a substantial performance boost ($+8.2\%$ at average mAP) when adding visual modalities, indicating the critical role of both modalities for the SED task.

{\bf Importance of motion information.}
In Table~\ref{subtab:kinetics}, we compare the performances of our models using different visual encoders. For both single-task (ST) and all-task (AT) models, the performances improve by a large margin when using the visual encoder~\cite{wang2023one} fine-tuned on Kinetics-400~\cite{carreira2017quo}. In particular, for TAL, the improvements of $3.4\%$ and $3.5\%$ at average mAP can be observed for the ST and AT models, respectively. This emphasizes the importance of motion information for video localization tasks, especially for the TAL task.

\begin{figure*}[t]
  \centering
  \setlength{\abovecaptionskip}{0.5mm}
   \includegraphics[width=1.0\linewidth]{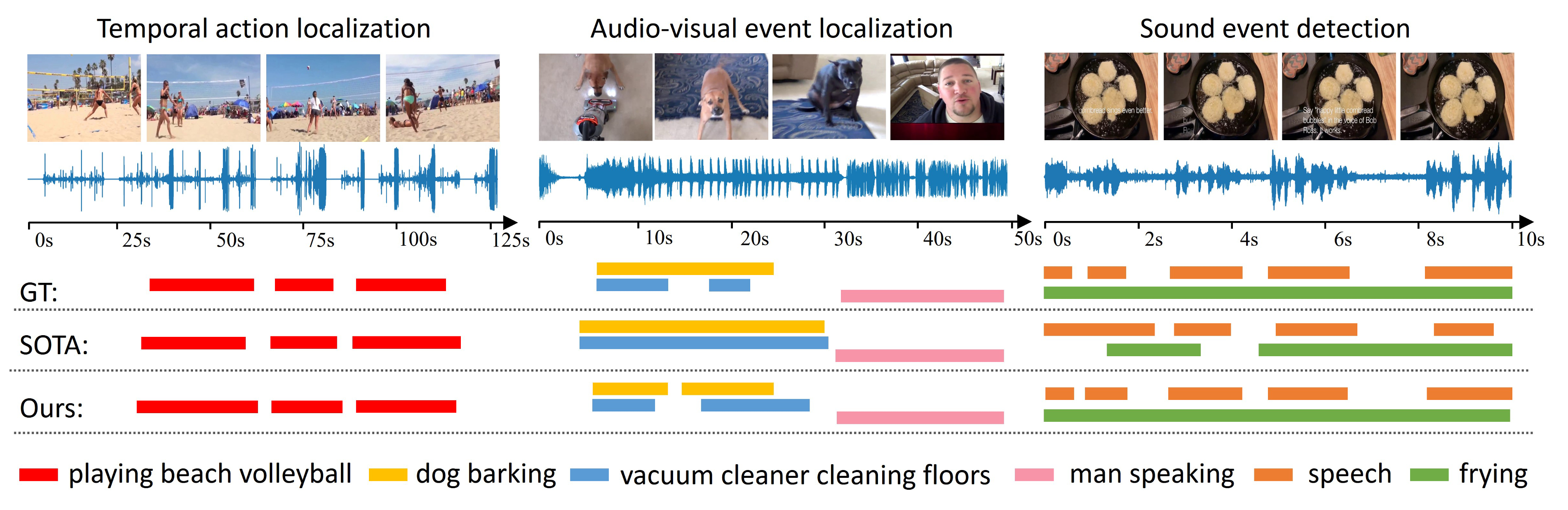}
   \vspace{-4mm}
   \caption{Qualitative results on TAL, AVEL and SED tasks. The examples are from the validation set of ActivityNet 1.3, the test set of UnAV-100, and the public evaluation set of DESED, respectively. ``GT'' is short for ground truth. ``SOTA'' denotes the state-of-the-art methods 
   (TriDet~\cite{shi2023tridet} for TAL, UnAV~\cite{geng2023dense} for AVEL, 
   and the audio-only single-task model for 
   SED, where all models use the same features). ``Ours'' is our AT model.
   We show the boundaries with the highest overlap with the ground truth.}
   \label{fig:vis1}
   \vspace{-2mm}
\end{figure*}

{\bf Comparison with other large-scale audio and visual encoders.}
In Table~\ref{tab:encoders}, we also explore the performances of other large-scale audio and visual encoders.
We show the results of the single-task model for the SED task. 
We find that using the audio embeddings extracted from CLAP~\cite{wu2023large} leads to poor performance. This may be attributed to its emphasis on global representations and insufficient fine-grained audio information modeling. Besides, CLIP~\cite{radford2021learning}, as a powerful visual-language pre-trained model, has comparable performances with other tri-modal aligned encoders. Moreover, LanguageBind~\cite{zhu2023languagebind} outperforms ImageBind~\cite{girdhar2023imagebind} in audio feature representation but is slightly weaker in visual features.
When using both audio and visual modalities, the ONE-PEACE features can achieve the best results.

\begin{figure}[!t]
  \centering
  \setlength{\abovecaptionskip}{0.5mm}
   \includegraphics[width=0.95\linewidth]{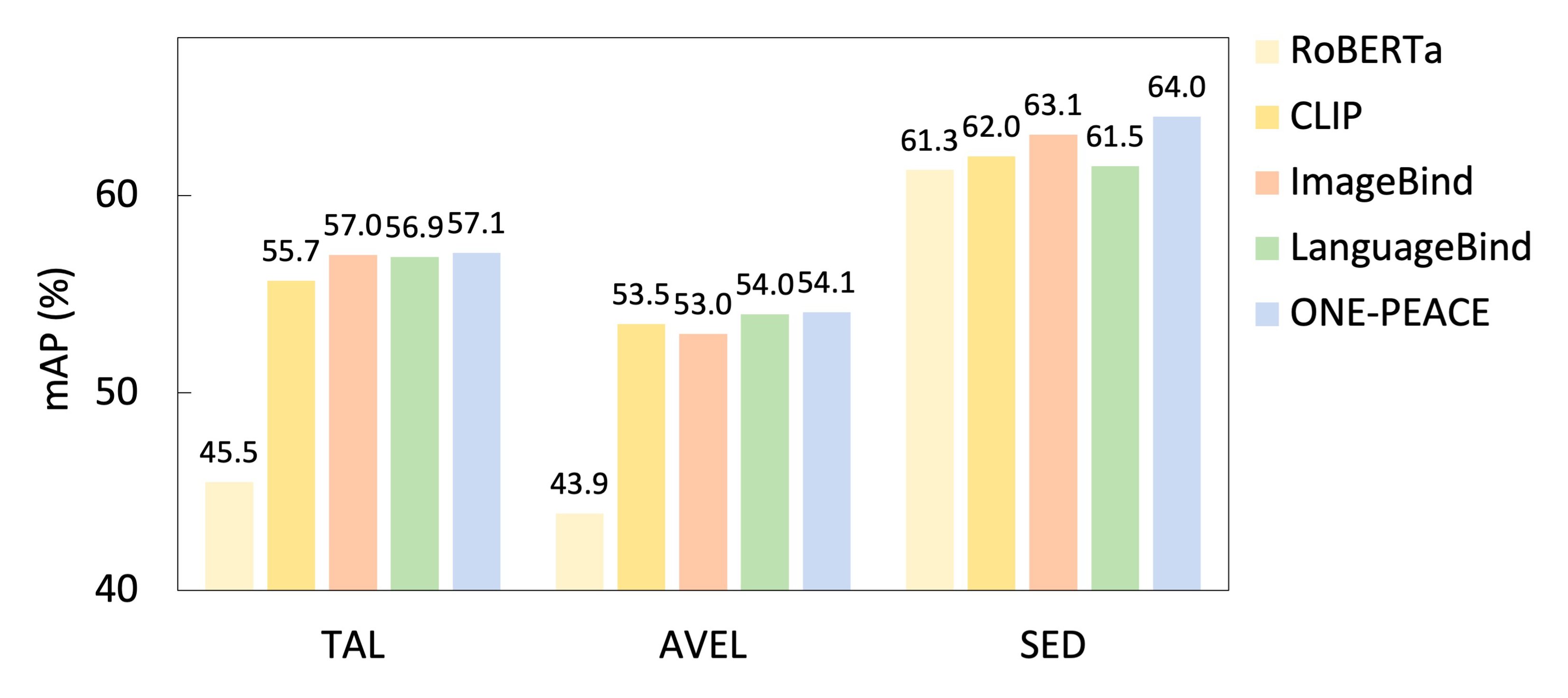}
   \caption{Localization results of our AT model on all three tasks with different text encoders for category embedding, measured by mAP@tIoU=0.5.}
   \label{fig:figtext}
   \vspace{-2mm}
\end{figure}

{\bf Comparison with other text encoders.}
In Fig.~\ref{fig:figtext}, we compare the performances of our all-task (AT) model using different text encoders for category embedding in the unified language-aware classifier. All models apply the audio and visual encoders of ONE-PEACE~\cite{wang2023one} for video representations by default. 
We can see that RoBERTa~\cite{liu2019roberta}, a large natural language processing (NLP) model, yields poor results across all three tasks, while due to the exceptional text encoding capability of CLIP, significant performance boosts can be observed. Besides, ImageBind~\cite{girdhar2023imagebind}’s text encoder performs better than that of LanguageBind~\cite{zhu2023languagebind} on TAL and SED tasks.
Furthermore, when utilizing the text encoder of ONE-PEACE, our AT model achieves the best results on all three tasks.

\begin{figure*}[t]
  \centering
  \setlength{\abovecaptionskip}{0.5mm}
   \includegraphics[width=1.0\linewidth]{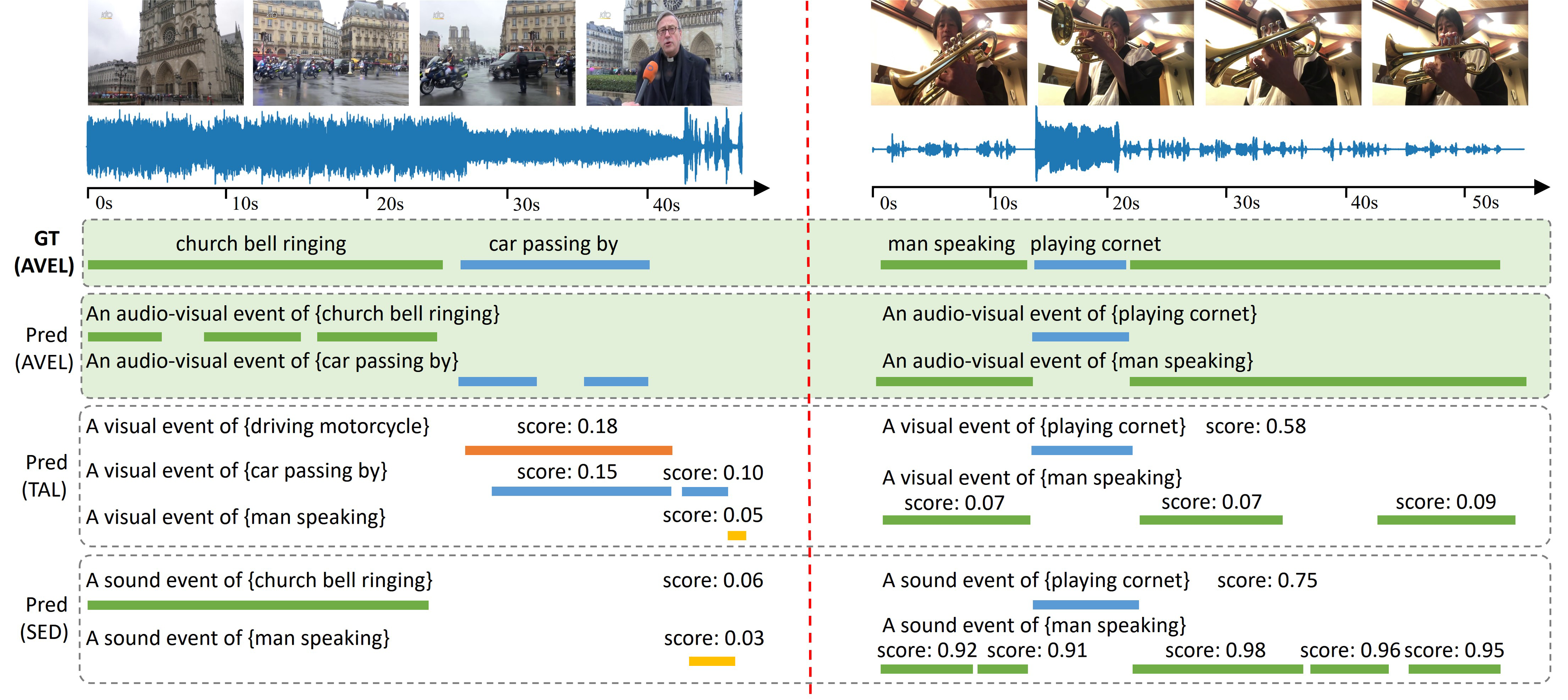}
   \caption{Examples of localizing the visual/sound events that are not present in ActivityNet 1.3/DESED. 
   The videos are from the test set of UnAV-100. ``GT (AVEL)'' denotes only AVEL annotations provided during training. ``Pred'' is the prediction results on each task. ``score'' is the confidence score of predictions. 
   }
   \label{fig:vis2}
\end{figure*}

\begin{figure}[!t]
  \centering
  \setlength{\abovecaptionskip}{0.5mm}
   \includegraphics[width=0.95\linewidth]{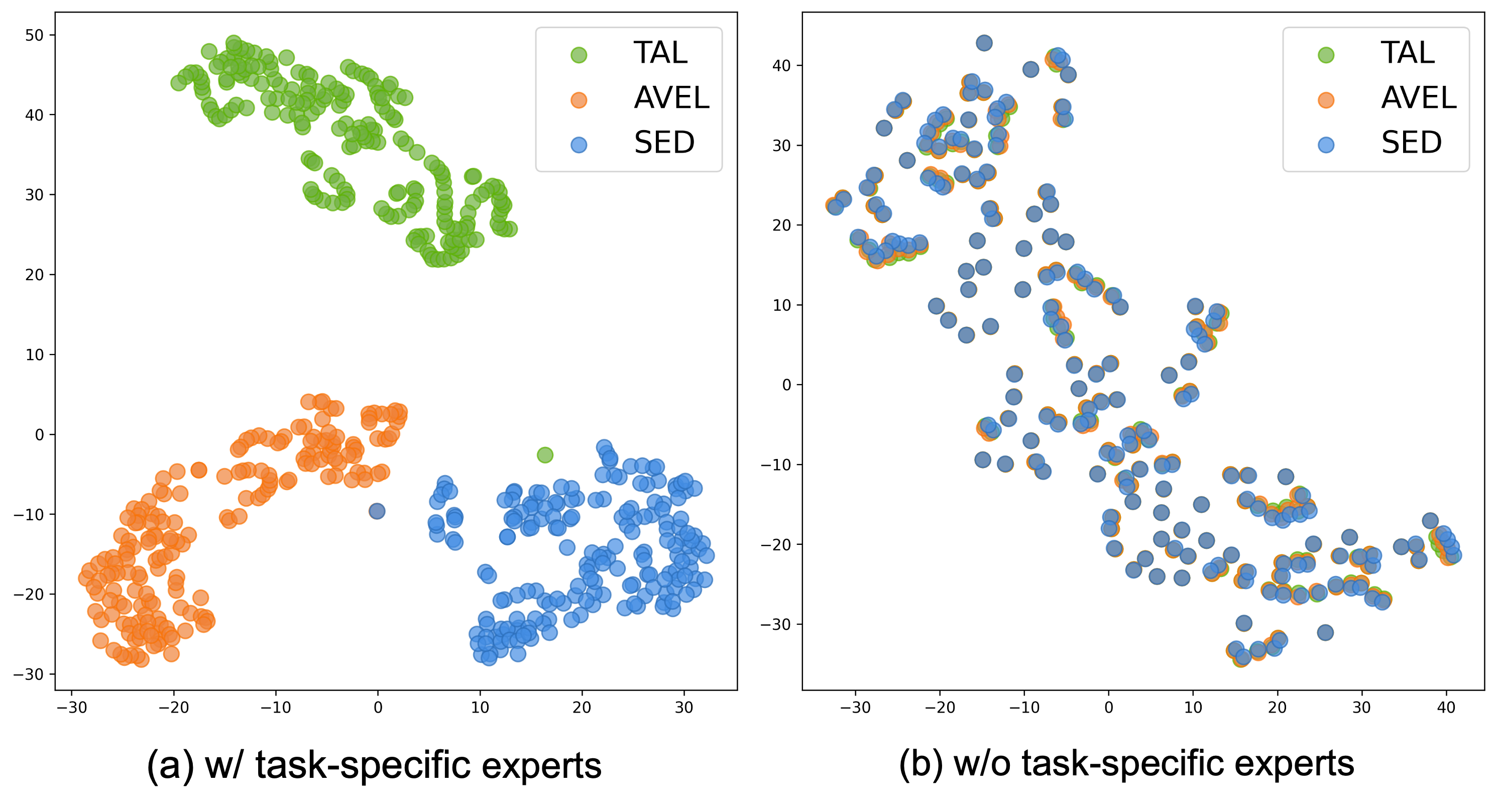}
   \vspace{-2mm}
   \caption{t-SNE visualization of the pyramid feature $Z^{1}$ learned by (a) UniAV with task-specific experts and (b) the multi-task variant without task-specific experts. Different colors denote different tasks, and each point denotes the feature of a temporal segment from the given video.}
   \label{fig:expert}
   \vspace{-2mm}
\end{figure}

\begin{figure}[!t]
  \centering
  \setlength{\abovecaptionskip}{0.5mm}
   \includegraphics[width=0.95\linewidth]{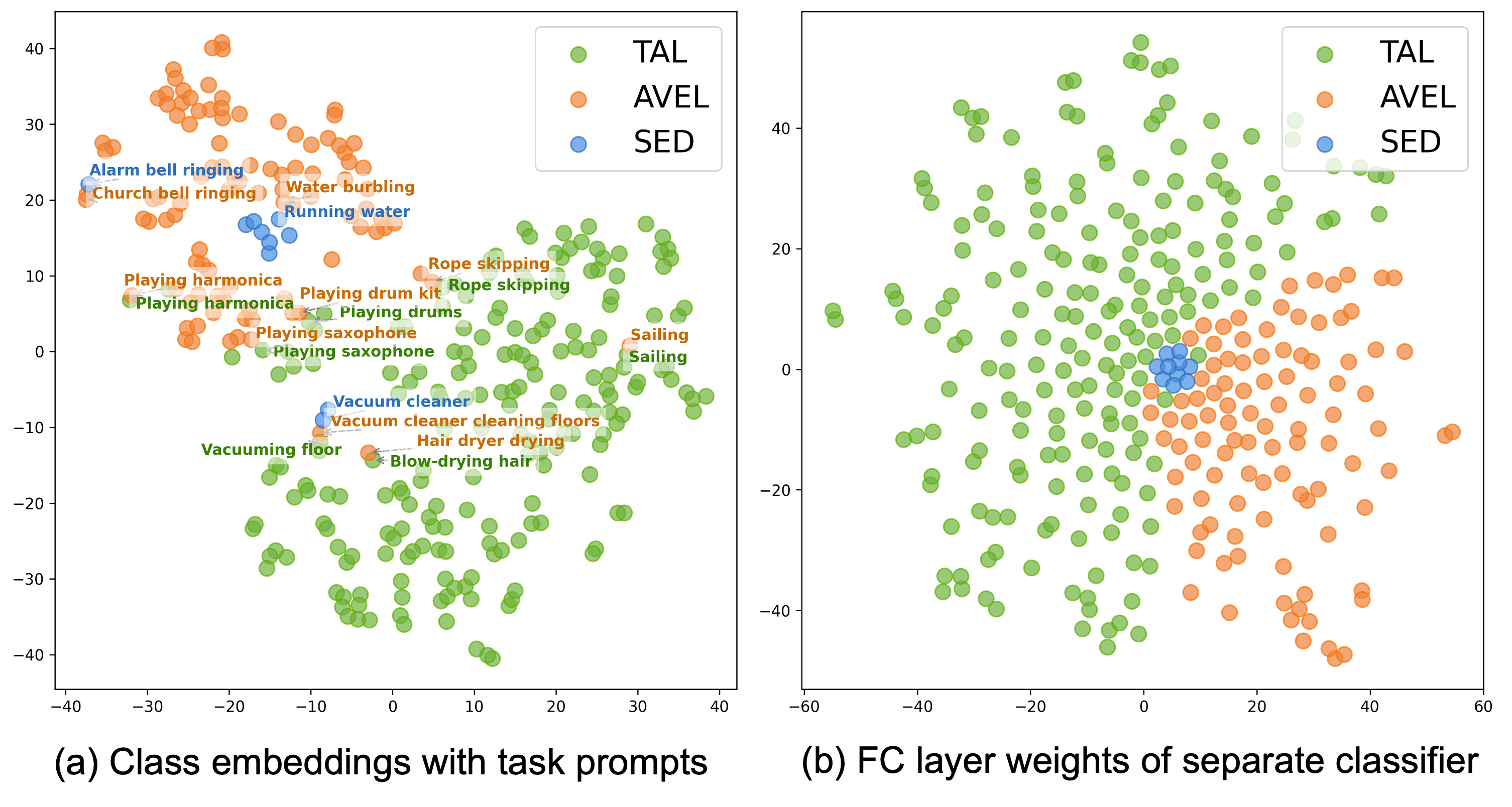}
   \vspace{-2mm}
   \caption{t-SNE visualization of (a) the class embeddings with task prompts encoded by ONE-PEACE text encoder, and (b) the last FC layer weights from the separate task-specific classification heads. Different colors denote different tasks, and each point denotes an embedding of an instance category.}
   \label{fig:text}
   \vspace{-2mm}
\end{figure}

\subsection{Visualization and Discussion}
{\bf Qualitative results.}
In Fig.~\ref{fig:vis1}, we visualize the predictions of our AT model on three tasks. Compared with the other state-of-the-art methods, our model generates localizations that better overlap with the ground truth. For instance, our model outputs finer boundaries to distinguish repeated occurrences of the same events, \eg, the audio-visual event of ``vacuum cleaner cleaning floors'' and the sound event of ``speech''. 
For SED task, visual information helps the model to detect the sound event of ``frying'' more accurately compared to that using audio modality alone. 

{\bf t-SNE visualization for TE and LCH.}
We first randomly sample a video for inference and visualize the output temporal feature sequences from the audio-visual pyramid transformer (level 1).
As shown in Fig.~\ref{fig:expert} (a), using task-specific experts (TE), the features for different tasks are clearly distinguishable, while without TE (Fig.~\ref{fig:expert} (b)), the features almost overlap. This reveals that TE enables the model to effectively learn distinct knowledge for each task.
Next, we visualize the class embeddings encoded by ONE-PEACE for unified classification (Eq.6). As in Fig.~\ref{fig:text} (a), the embeddings reveal rich correlations between event categories across tasks, \eg, ``a visual event of \{playing drums\}" and ``an audio-visual event of \{playing drum kit\}". In contrast, the last FC layer weights of separate task-specific classification heads show no inter-task correlations (Fig.~\ref{fig:text} (b)). It suggests that our LCH breaks the constraint of fixed, independent class sets of different tasks, effectively fostering semantic associations between events across tasks.

{\bf Localizing instance categories across tasks.}
Benefiting from the language-aware classification head that utilizes the pre-trained text encoder~\cite{wang2023one}, our AT model has a novel capability of localizing the categories from other task datasets when performing the current task. 
For example, we select some categories from the UnAV-100 dataset that are not present in ActivityNet 1.3 and DESED, and then tokenize them with the TAL and SED prompts and concatenate with the original class embeddings to conduct inference on all three tasks.
In Fig.~\ref{fig:vis2}, we observe that our AT model successfully detects the visual events of ``driving motorcycle'', ``car passing by'', and ``playing cornet'', \etc, and the sound events of ``church bell ringing'' and ``man speaking", \etc, even though the model just learned the audio-visual events of these categories during training.
It indicates that our model has the remarkable flexibility to localize rich instance categories across tasks by simply changing task prompts, greatly alleviating the data scarcity for each task, especially for SED. 
UniAV is capable of detecting all three types of instances (visual, sound and audio-visual) for the categories of all three datasets (total 310 classes).

\begin{figure*}[t]
  \centering
  \setlength{\abovecaptionskip}{0.5mm}
   \includegraphics[width=1.0\linewidth]{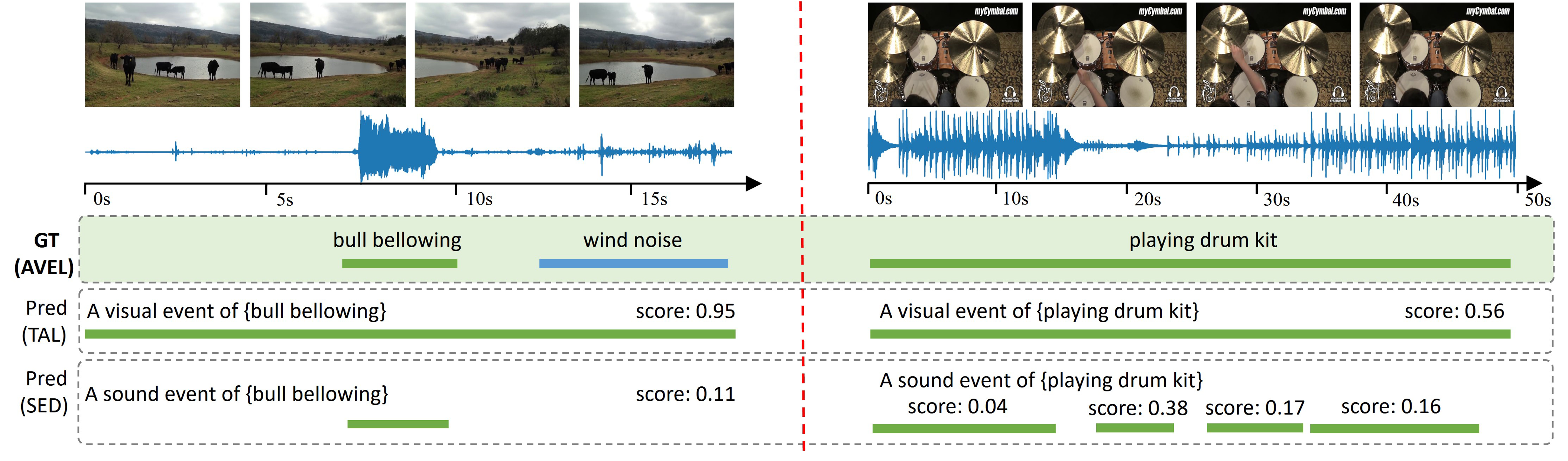}
   \caption{Examples of open-vocabulary localization on UnAV-100. 
   The multi-task model trained on ActivityNet 1.3 for TAL and DESED for SED is utilized to detect unseen instance categories from UnAV-100. The videos are from the test set of UnAV-100.}
   \label{fig:vis3}
\end{figure*}

\begin{figure*}[t]
  \centering
  \setlength{\abovecaptionskip}{0.5mm}
   \includegraphics[width=1.0\linewidth]{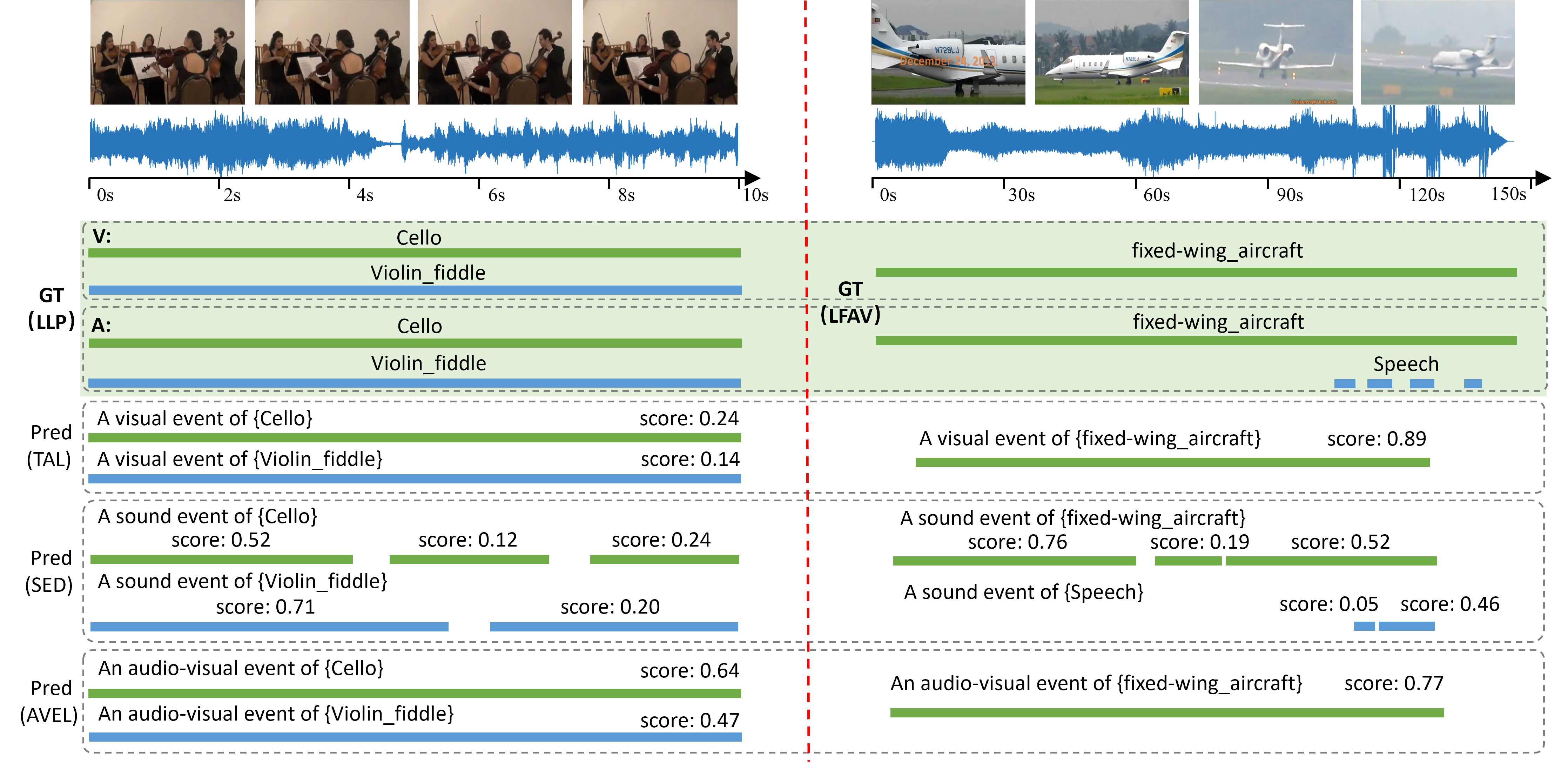}
   \caption{Examples of open-vocabulary localization on LLP~\cite{tian2020unified} and LFAV~\cite{hou2024toward} using our UniAV trained on ActivityNet 1.3 (TAL), DESED (SED) and UnAV (AVEL). The videos are from the LLP (left) and LFAV (right) test sets, respectively. ``GT" is the LLP/LFAV annotations of visual (V) and audio (A) events.}
   \label{fig:zero-shot}
\end{figure*}

{\bf Emergent open-vocabulary localization.}
Apart from significant gains for closed-set performances, our model also showcases an impressive ability of open-set
localization for novel categories, owing to the generalizability of the used tri-modal model~\cite{wang2023one} and the unified language-aware classification head.
First, as shown in Fig.~\ref{fig:vis3}, there is no class ``bull bellowing'' and ``playing drum kit'', even a similar one in ActivityNet 1.3 and DESED category sets, but we surprisingly found that the multi-task model (row 4 in Table~\ref{tab:multitask}) can accurately detect the corresponding visual and sound events with high confidence scores.
Additionally, we also conduct open-set localization on LLP~\cite{tian2020unified} and LFAV~\cite{hou2024toward} datasets. 
In Fig.~\ref{fig:zero-shot}, although our training data does not contain the categories ``cello", ``Violin\_fiddle", and ``fixed-wing\_aircraft" (only the similar ones like ``playing cello/violin" and ``airplane flyby"). 
Despite this, our model successfully detects the corresponding visual/sound/audio-visual events with high confidence.  
It demonstrates our model's strong potential in open-set capabilities, greatly distinguishing it from other approaches.

\section{Conclusion}
This paper proposes a Unified Audio-Visual perception network, UniAV, for the joint learning of TAL, SED and AVEL tasks for the first time, realizing the localization of visual actions, sound events and audio-visual events in an untrimmed video by a single unified model. 
Specifically, we introduce a unified audio-visual encoding pipeline to minimize data discrepancies across tasks, and meanwhile applying task-specific experts to capture distinct knowledge for each task. Moreover, a novel unified language-aware classifier allows the model to have high flexibility and generalizability during inference.
Extensive experiments demonstrate that our UniAV achieves superior and competitive performances on three challenging benchmarks, surpassing its single-task counterparts and the naive multi-task baseline using only a single model parameters.
In the future, we plan to utilize more available data and fully leverage existing large-scale multi-modal pre-trained models to further explore the model's capabilities in open-world predictions.

\bibliographystyle{IEEEtran}
\bibliography{main}

\vfill

\end{document}